%% file: 1.main.tex
\documentclass{article}
\usepackage{arxiv}

\usepackage[utf8]{inputenc} 
\usepackage[T1]{fontenc}    
\usepackage{hyperref}       
\usepackage{url}            
\usepackage{booktabs}       
\usepackage{nicefrac}       
\usepackage{microtype}      
\usepackage{lipsum}		
\usepackage[numbers,sort&compress]{natbib}
\usepackage{doi}
\usepackage{algorithm}
\usepackage{algorithmic}
\usepackage{amsfonts,amsmath,amssymb,amsthm}
\usepackage{multirow}
\usepackage{threeparttable}
\usepackage[capitalise]{cleveref}
\usepackage{textcomp}
\usepackage{comment}
\usepackage{subcaption}
\usepackage{multicol}
\usepackage{fixltx2e}
\usepackage{relsize}
\usepackage{mathbbol}
\usepackage{xurl}
\usepackage{wasysym}
\usepackage{bm}
\usepackage{graphicx}
\usepackage{xcolor}
\usepackage{tcolorbox}
\usepackage{enumitem}

\usepackage{mdframed}
\newmdenv[
  linewidth=1pt, 
  roundcorner=5pt, 
  linecolor=black, 
  backgroundcolor=white, 
  innerleftmargin=10pt, 
  innerrightmargin=10pt, 
  innertopmargin=10pt, 
  innerbottommargin=10pt 
]{custombox}

\newcommand\edit[1]{\textcolor{black}{#1}}

\title{A RAG-Based Multi-Agent LLM System for Natural Hazard Resilience and Adaptation}

\author{
    Yangxinyu Xie \thanks{Department of Statistics and Data Science, University of Pennsylvania, Philadelphia, PA. \texttt{xinyux@wharton.upenn.edu}} \And Bowen Jiang \thanks{Department of Computer and Information Science, University of Pennsylvania, Philadelphia, PA.} \And 
    Tanwi Mallick \thanks{Mathematics and Computer Science Division, Argonne National Laboratory, Lemont, IL. \texttt{tmallick@anl.gov
    }} \And
    Joshua Bergerson
    \thanks{Decision and Infrastructure Sciences Division, Argonne National Laboratory, Lemont, IL. }
    \And
    John K. Hutchison $^{\S}$ \And
    Duane R. Verner $^{\S}$ \And
    Jordan Branham $^{\S}$ \And
    M. Ross Alexander $^{\S}$ \And
    Robert B. Ross $^{\ddagger}$ \And
    Yan Feng \thanks{Environmental Science Division, Argonne National Laboratory, Lemont, IL.}
    \And
    Leslie-Anne Levy $^{\S}$ \And
    Weijie~J. Su $^{\ast}$ \And Camillo J. Taylor $^{\dagger}$
}

\hypersetup{
pdftitle={A RAG-Based Multi-Agent LLM System for Natural Hazard Resilience and Adaptation}
pdfsubject={Large Language Models, Natural Hazards, Extreme Weather Events},
pdfkeywords={Large Language Models, Natural Hazards, Extreme Weather Events}
}

\begin{document}

\maketitle

\begin{abstract}
Large language models (LLMs) are a transformational capability at the frontier of artificial intelligence and machine learning that can support decision-makers in addressing pressing societal challenges such as extreme natural hazard events. As generalized models, LLMs often struggle to provide context-specific information, particularly in areas requiring specialized knowledge. 
\edit{In this work we propose a retrieval-augmented generation (RAG)-based multi-agent LLM system to support analysis and decision-making in the context of natural hazards and extreme weather events. As a proof of concept, we present WildfireGPT, a specialized system focused on wildfire hazards. The architecture employs a user-centered, multi-agent design to deliver tailored risk insights across diverse stakeholder groups. By integrating natural hazard and extreme weather projection data, observational datasets, and scientific literature through an RAG framework, the system ensures both the accuracy and contextual relevance of the information it provides.
Evaluation across ten expert-led case studies demonstrates that WildfireGPT significantly outperforms existing LLM-based solutions for decision support.
}
\end{abstract}

\section*{Introduction}
\input{2.intro}

\section*{Results} 
\input{3.results}

\section*{Discussion} 
\input{4.discussion}

\section*{Methodology}
\input{7.method}

\section*{Data Availability}
\input{5.data}

\section*{Code Availability}
\input{6.code}

\section*{Acknowledgment}
This material is based upon work supported by Laboratory Directed Research and Development funding from Argonne National Laboratory, provided by the Director, Office of Science, of the U.S. Department of Energy under Contract No. DEAC02-06CH11357. Y. Xie and W. J. Su acknowledge support from the NSF HDR TRIPODS award (CCF-1934876). B. Jiang and C. J. Taylor acknowledge support from the NSF grant CCF-2112665 (TILOS). M. R. Alexander, J. Branham, and Y. Feng contributed significantly to the case studies. The authors also thank Jane S. Frantz, Parfait Gasana, Vijay P. Ramalingam, and Charles Doktycz for their valuable contributions to the case studies. The funder played no role in study design, data collection, analysis and interpretation of data, or the writing of this manuscript. 

\bibliographystyle{unsrt}
\bibliography{bibfile, nature}

\input{8.appendix}
\end{document}

%% file: 2.intro.tex
\edit{Natural hazards and extreme weather events such as wildfires, floods, and hurricanes present significant operational and management challenges across sectors, particularly in the management of critical infrastructure systems \citep{tabari2020climate, mora2022over, national2016attribution, task2021impacts, lu2024oxygenerator, IPCC_AR6_2021}. 
From intensifying heatwaves \citep{lopez2018early} and floods \citep{hirabayashi2013global} to more frequent and severe wildfires \citep{senande2022spatial, brown2023climate}, natural hazard events are becoming increasingly disruptive to infrastructure networks, demanding more effective tools for risk assessment, planning, and response.}

Although scientific research provides the building blocks for a wide range of solutions to address the \edit{increasing risk from natural hazards}, the complexity and volume of \edit{published} research often hinder the efficient translation of scientific knowledge into \edit{risk reduction policies and programs} \citep{adshead2023mission}. This disconnect is particularly problematic for professionals \edit{at the forefront of managing hazard-related risks,} such as urban planners, emergency managers, and infrastructure operators, who may have \edit{limited} access to the latest knowledge and methods to mitigate \edit{these threats}. Recent advances in natural language processing, especially large language models (LLMs), present an innovative solution for democratizing \edit{science on natural hazard resilience} and facilitating knowledge transfer \citep{callaghan2021machine, webersinke2022climatebert, planas2022beyond, rolnick2022tackling, spokoyny2023towards, schimanski2023climatebert, schimanski2024bridging}. LLMs possess the potential to process and synthesize vast amounts of textual information and explain them through conversations, making crucial information accessible to people from diverse backgrounds \citep{rillig2023risks, bulian2023assessing, chen2023foundation, toetzke2023leveraging, stagnol2023answering, atkins2024generative, muccione2024integrating, hsu2024evaluating}. 

\edit{Recent research has explored the capabilities of LLMs in natural hazard science and management contexts. DisasterResponseGPT enables users to input disaster scenario descriptions and receive action plans as outputs \citep{goecks2023disasterresponsegpt}. To explore the potential for LLMs to revolutionize fire engineering, researchers evaluated LLM performance across several fire engineering scenarios, including structural design, prevention strategies, and regulatory compliance \citep{hostetter2024large}. Building on these evaluations, researchers have explored multi-round prompting techniques that allow users to iteratively refine LLM responses with additional context, enhancing disaster management applications \citep{goecks2023disasterresponsegpt, chen2024optimizing}. To address limitations in contextual understanding and domain-specific knowledge, ChatClimate \citep{vaghefi2023chatclimate} and ClimateGPT \cite{thulke2024climategpt} incorporate assessment reports from the Intergovernmental Panel on Climate Change (IPCC). More sophisticated implementations connect LLMs with climate models such as the Model for the Assessment of Greenhouse Gas Induced Climate Change \citep{cao2024llm}. In the context of flood risk management, researchers have integrated relevant geospatial and demographic data \citep{martelo2024towards}. Beyond academic solutions, commercial tools such as Perplexity AI
\cite{perplexity2024} offer LLM-integrated search capabilities that can retrieve and synthesize relevant hazard information from current scientific literature. These advancements demonstrate the potential of LLMs to bridge the gap between scientific research and practical natural hazard risk management applications.}

Nonetheless, existing LLM research for \edit{natural hazard} decision support faces fundamental limitations in personalization, data utilization, and evaluation. First, prior studies primarily focus on LLMs' performance on generic, one-off questions \citep{vaghefi2023chatclimate, thulke2024climategpt, pirozelli2024benchmarks, hsu2024evaluating, cao2024llm, hostetter2024large, martelo2024towards, chen2024optimizing, goecks2023disasterresponsegpt}. This approach assumes users possess sufficient expertise to articulate clear queries, disregarding the complex reality that professionals have varying backgrounds and levels of expertise and often require multiple rounds of interaction to capture their context-specific information needs. This issue is  compounded by the tendency of current LLMs to provide hasty, generic responses without requesting \edit{clarifying} details \edit{necessary for developing detailed, context-specific responses} \citep{hager2024evaluation, li2024mediq}. The combination of these factors---assuming user expertise and providing quick, undifferentiated answers---can lead to significant misunderstandings or oversimplifications.  Second, \edit{the spatial heterogeneity of natural hazards and infrastructure vulnerability demands location-specific analyses at granular resolution. The} pronounced lack of textual (\edit{i.e., }scientific papers and reports) and data-based (\edit{i.e., }projections and observational data) knowledge \edit{integration} in existing research \citep{vaghefi2023chatclimate, toetzke2023leveraging, stagnol2023answering, hsu2024evaluating,thulke2024climategpt,kraus2023enhancing,cao2024llm} hinders the ability to \edit{synthesize} comprehensive evidence grounded in local contexts \citep{larosa2024artificial}. \edit{Our comparative analysis with existing alternatives such as ChatClimate \cite{vaghefi2023chatclimate} and Perplexity AI  \cite{perplexity2024} shows that standard LLM applications fail to reliably interpret complex grid-structured data formats, critically limiting their utility for location-specific risk assessment and infrastructure planning. Third, systematic evaluation of LLMs at the intersection of the natural hazard and infrastructure domains presents significant methodological challenges. Conventional evaluation frameworks based on natural language processing predominantly rely on metrics that quantify lexical similarity or assess linguistic fluency approaches that are inadequate for domain-specific applications requiring actionable insights. While some studies have explored fact-checking of LLM-generated content \citep{vaghefi2023chatclimate, hsu2024evaluating, bulian2023assessing}, such verification represents only a preliminary step toward comprehensive evaluation. Critically absent from current assessment protocols is the systematic measurement of response utility within decision-making contexts—specifically, the capacity of LLM outputs to provide contextually appropriate, implementable recommendations for infrastructure resilience that accurately reflect localized hazard conditions. This evaluation gap necessitates specialized expertise capable of assessing both factual accuracy and practical applicability within complex sociotechnical systems.}

To address these limitations, we introduce an innovative multi-agent LLM prototype designed as a co-pilot for understanding \edit{natural} hazards and developing adaptation strategies. Our approach encompasses several key contributions:

\begin{enumerate}
    \item Human-Centered Personalization: The agents are designed to engage users in a series of questions to understand their professional background, location of interest, and specific concerns about \edit{natural} hazards. Based on this input, the system develops a strategic plan for analysis, determining which datasets to examine, what literature to review, and the appropriate scope for recommendations. 
    \item Data Integration and Interactive Visualization: We combine location-specific projections, hazard history, census characteristics, and scientific literature to augment LLM responses \edit{using a retrieval-augmented generation (RAG) approach}. Moreover, our system integrates interactive visualizations of geospatial data with textual outputs, enabling nuanced user exploration of multidimensional risk factors.
    \item Three-Stage Evaluation: \edit{We assess the performance of  WildfireGPT through 10 case studies covering a diverse range of topics and locations. Our evaluation comprises three complementary stages: (1) a modular comparison against ChatClimate and Perplexity AI for data and literature retrieval effectiveness, (2) a qualitative ablation study examining how profile specificity impacts response quality, and (3) a detailed assessment of case study responses through fidelity verification, expert evaluation of utility, and exploration of LLM-as-a-judge for automated assessment} \cite{liu2023g, zheng2023judging}. 
    This \edit{evaluation process} focuses on practical value and real-world performance. By incorporating LLM-as-a-judge, we explore the potential for scalable, automated evaluation that could maintain quality assurance during deployment without constant human oversight.
\end{enumerate}

As a proof of concept we develop an LLM tool to support decision-making surrounding wildfire risk and resilience \edit{within the United States}. \edit{Projected environmental and land-use changes are expected} to alter the frequency, severity, seasonality, and spread of wildfires across the United States \citep{senande2022spatial, task2021impacts, prestemon2016projecting, mckenzie2017climate}.Projected increases in lightning frequency are expected to increase the frequency of wildfire ignitions \citep{romps2014projected, krause2014sensitivity, hessilt2022future}. At the same time, the frequency, duration, severity, and seasonality of fire weather conditions are projected to increase, including increasing severity and frequency of drought conditions due to shifting precipitation patterns, increasing ambient and extreme temperatures, and changing wind patterns \citep{richardson2022global, gamelin2022projected}. Beyond \edit{these environmental} impacts, population growth, increased development in and adjacent to wildfire-prone areas, and modern fire suppression policies are expected to further increase wildfire risk \citep{task2021impacts, pausas2021wildfires}. This paper documents the development, methodology, and initial findings of the WildfireGPT prototype through comprehensive case studies and evaluation, offering a new paradigm for artificial intelligence (AI)-assisted decision-making in \edit{natural hazard risk management}. 

%% file: 3.results.tex
\begin{figure}[hbt!]
    \begin{center}
    \includegraphics[width=\textwidth]{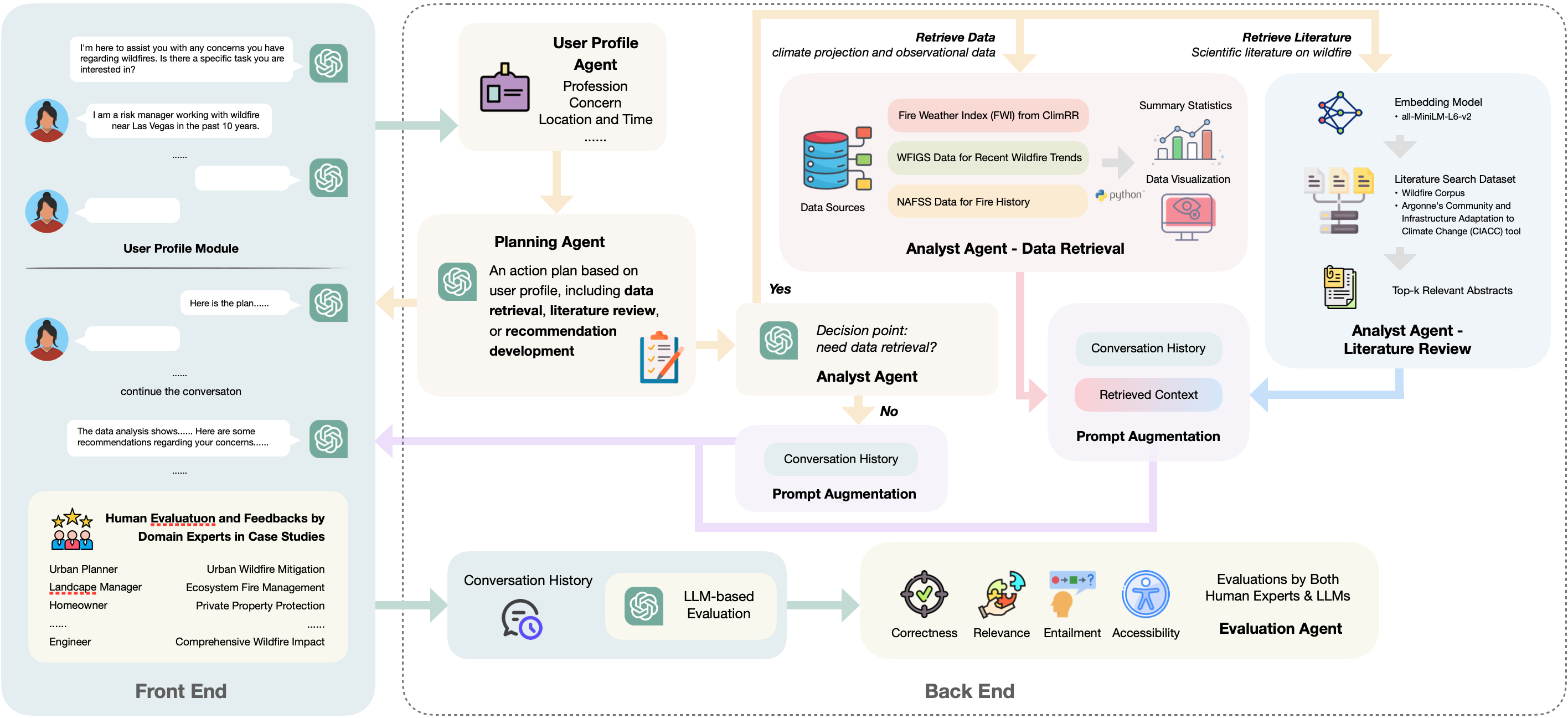}
    \end{center}
    \caption{Overview of WildfireGPT architecture comprising an LLM agent integrated with the multi-agent RAG framework. The WildfireGPT prototype focuses on enhancing consulting interactions using LLM agents stepping through a multistage approach. Its {\bf user profile agent} engages the user with a tailored questionnaire to create a detailed profile; the {\bf planning agent} formulates a customized action plan addressing the user's queries and concerns, ensuring alignment with their evolving needs; and the {\bf analyst agent} aids in accessing and interpreting relevant data and literature and providing recommendations.}
    \label{fig:overview}
\end{figure}

\begin{figure}[hbt!]
    \centering
    \includegraphics[width=0.32\linewidth]{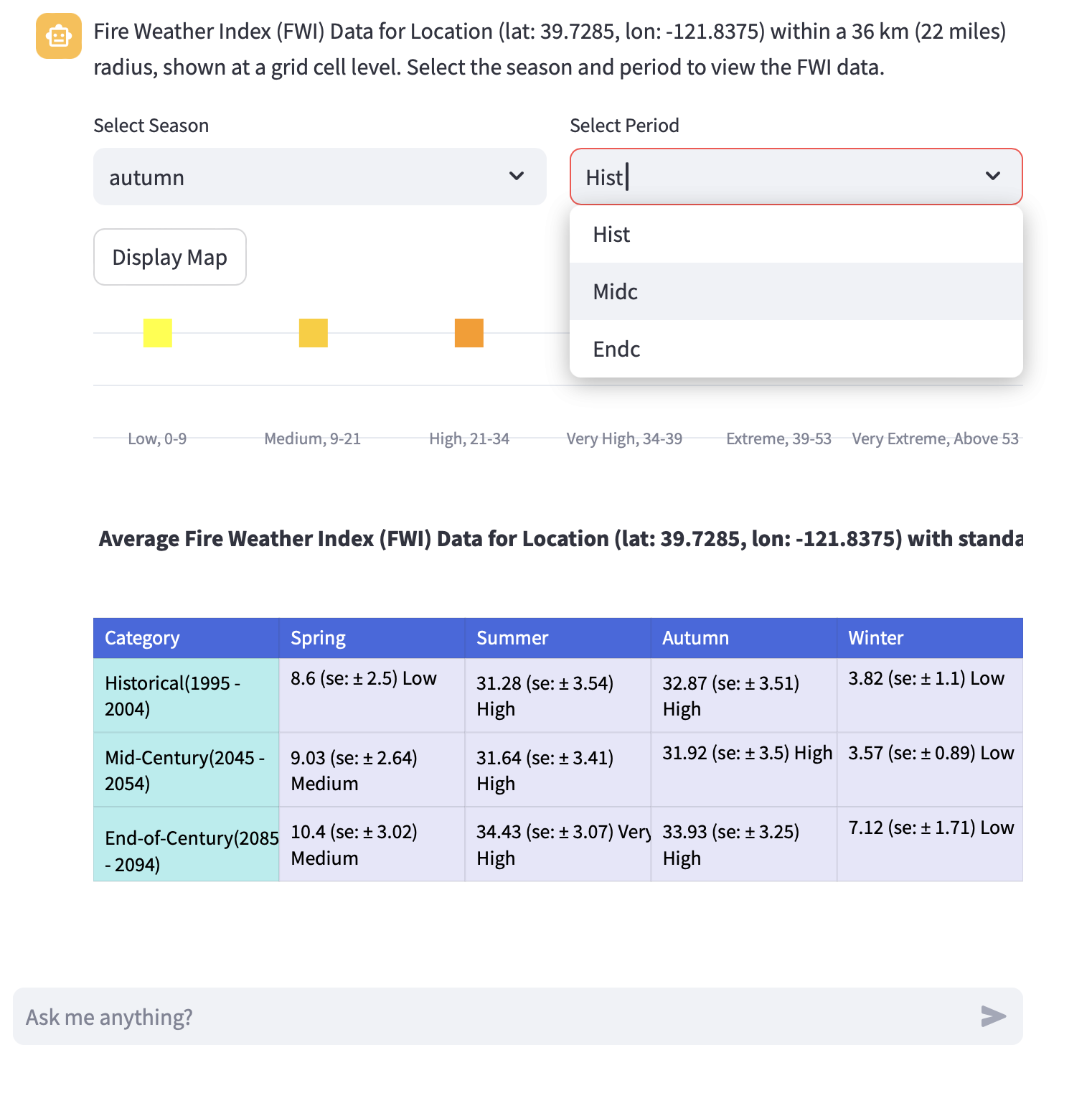}
    \includegraphics[width=0.32\linewidth]{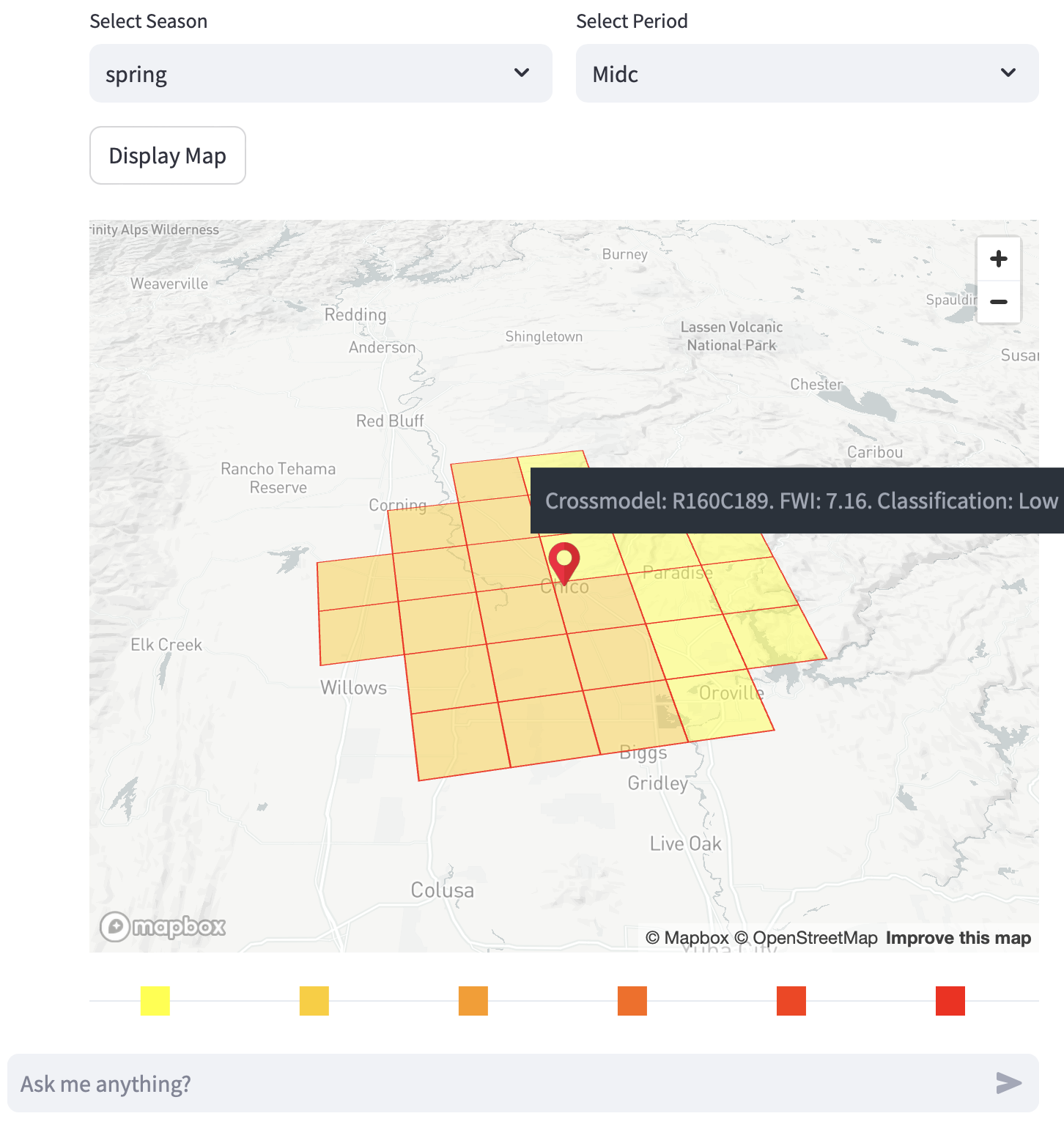}
    \includegraphics[width=0.32\linewidth]{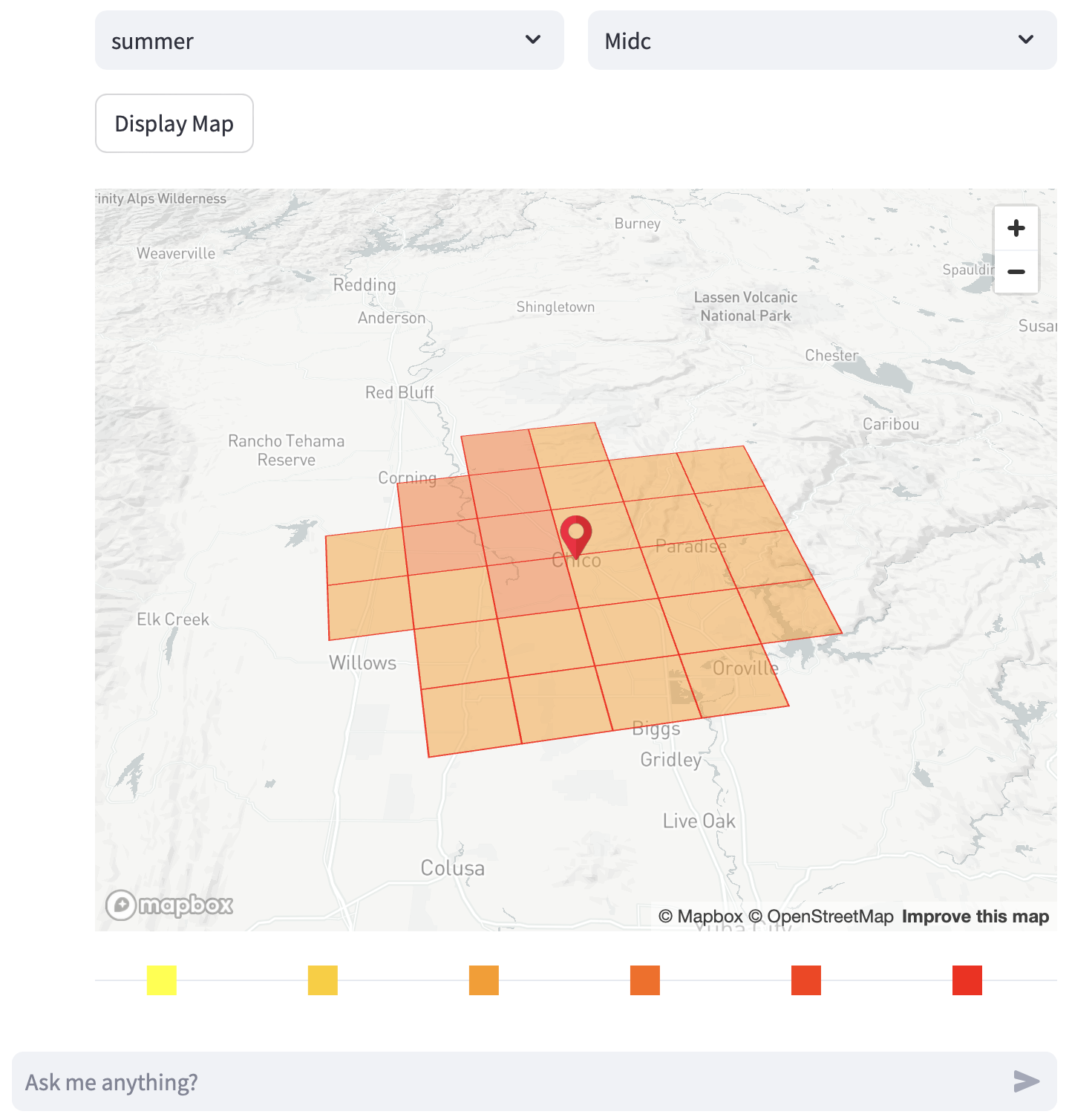}
    \includegraphics[width=0.32\linewidth]{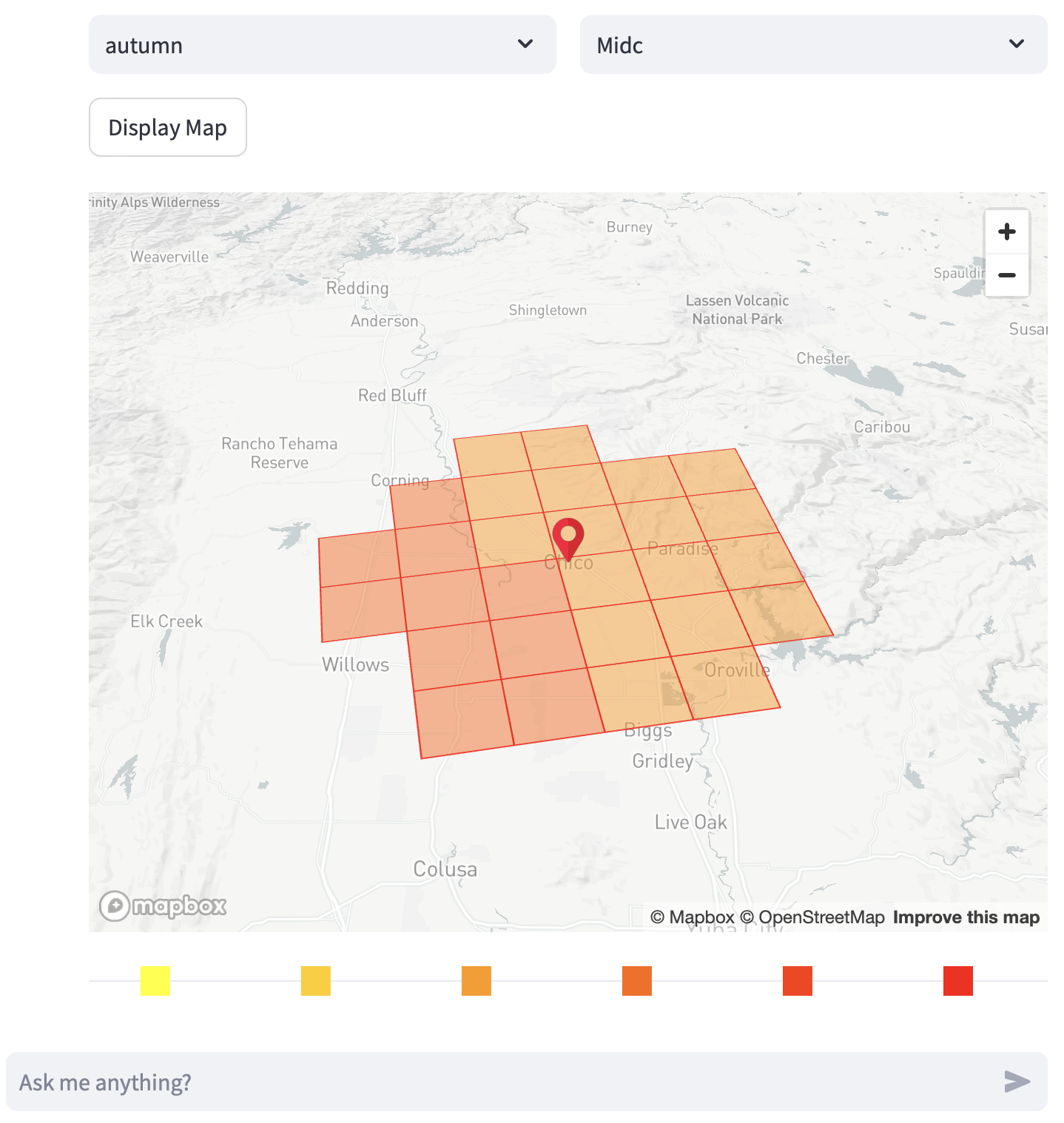}
    \includegraphics[width=0.32\linewidth]{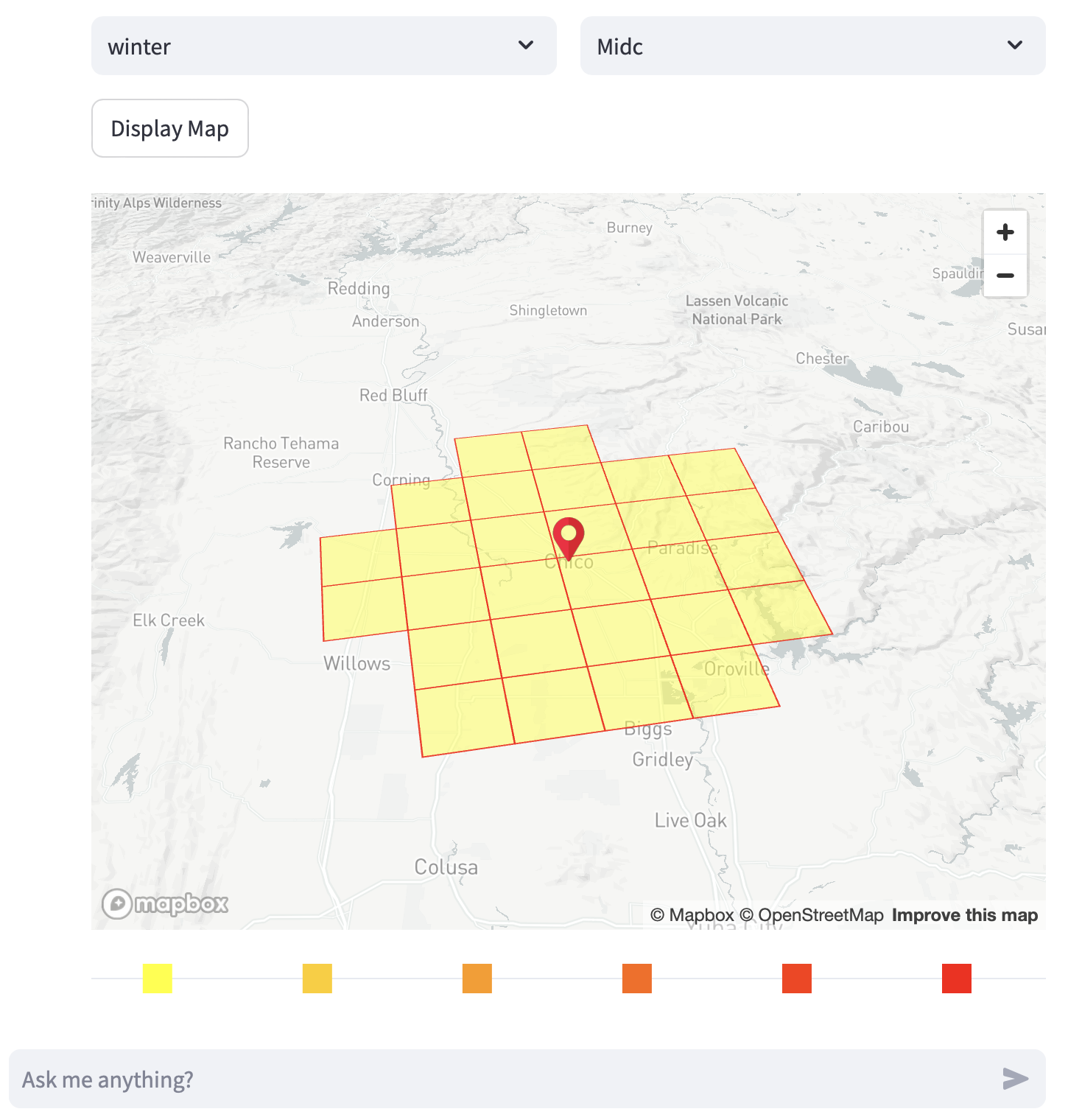}
    \includegraphics[width=0.32\linewidth]{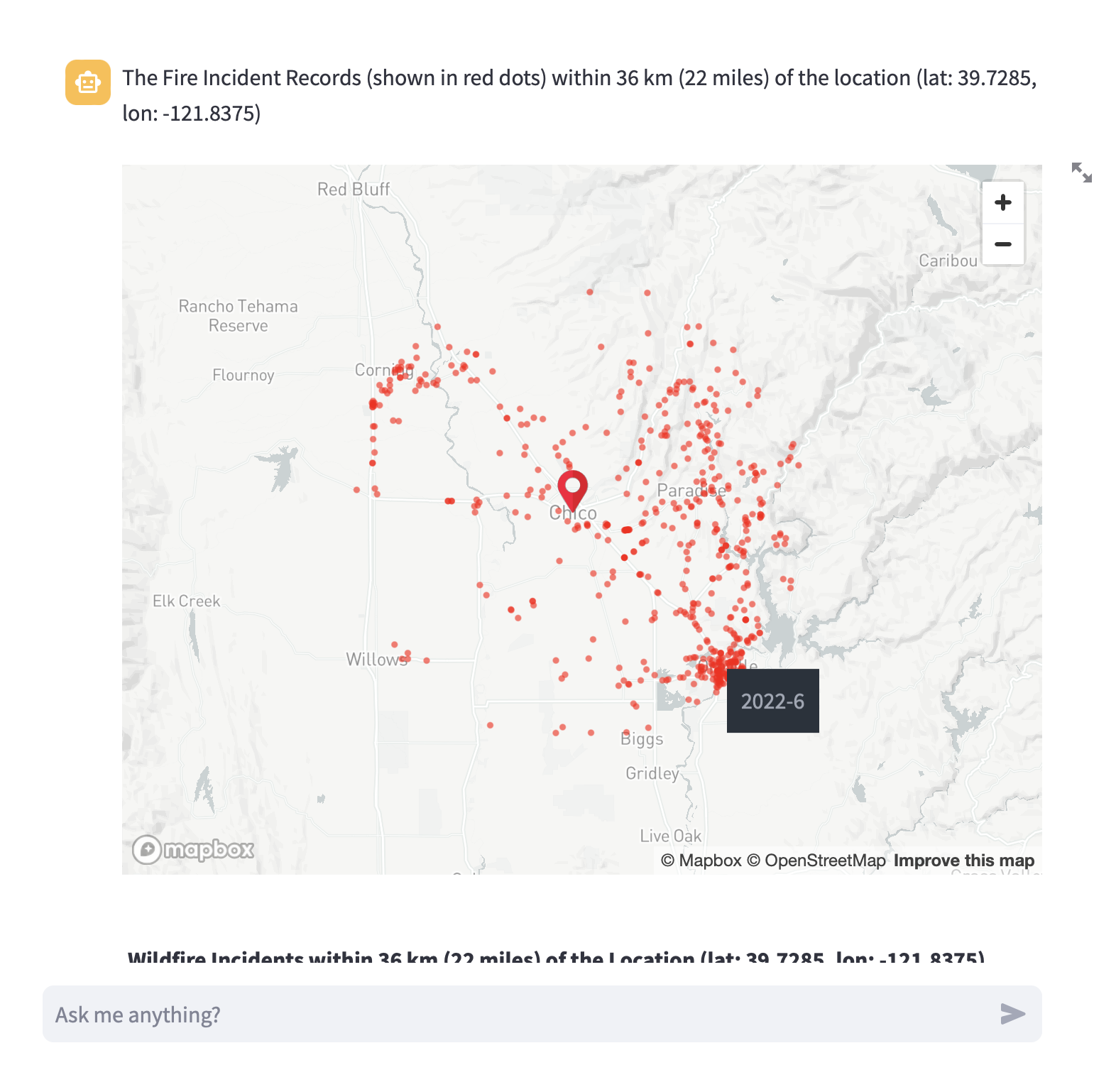}
    \includegraphics[width=0.32\linewidth]{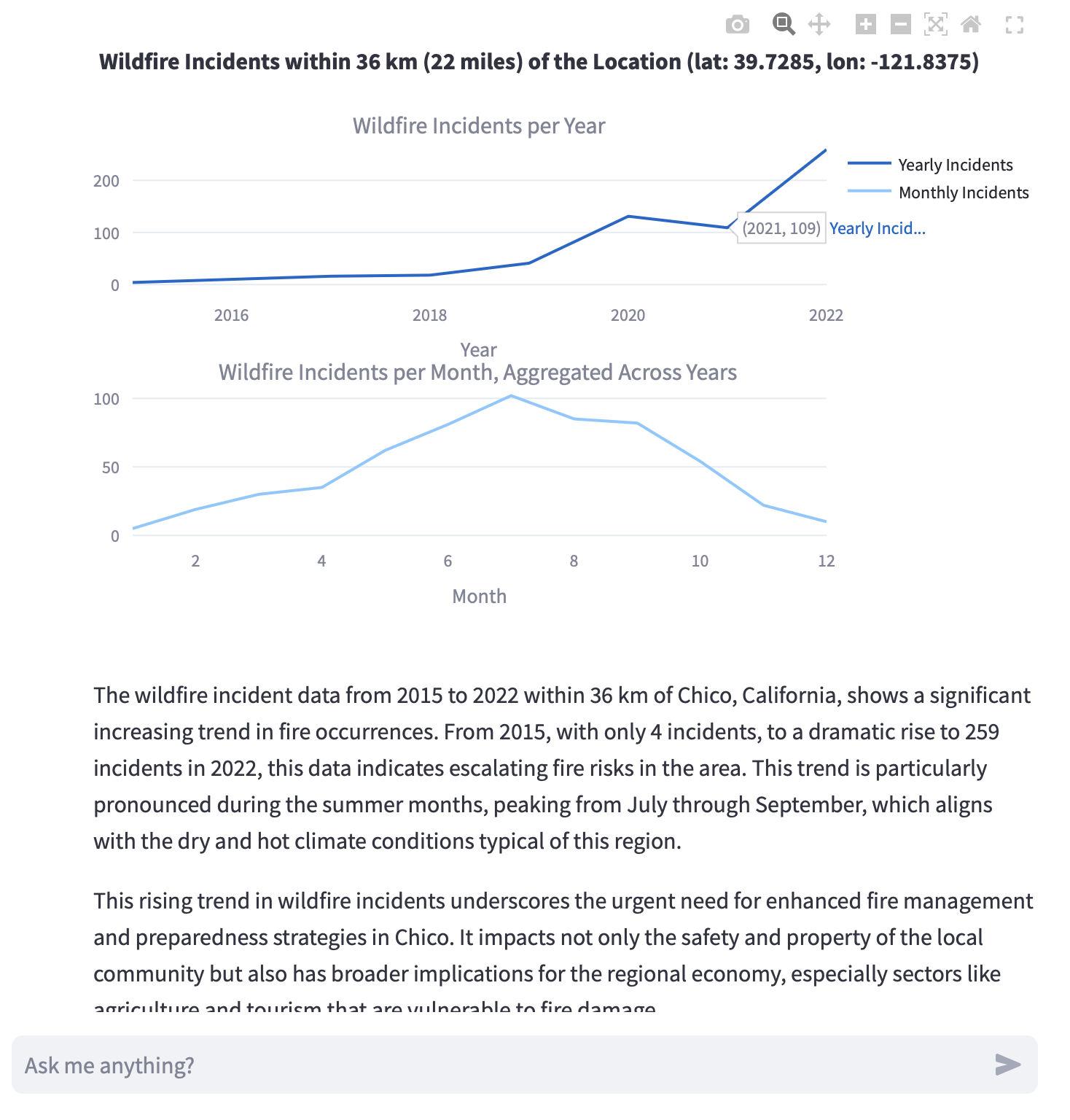}
    \includegraphics[width=0.32\linewidth]{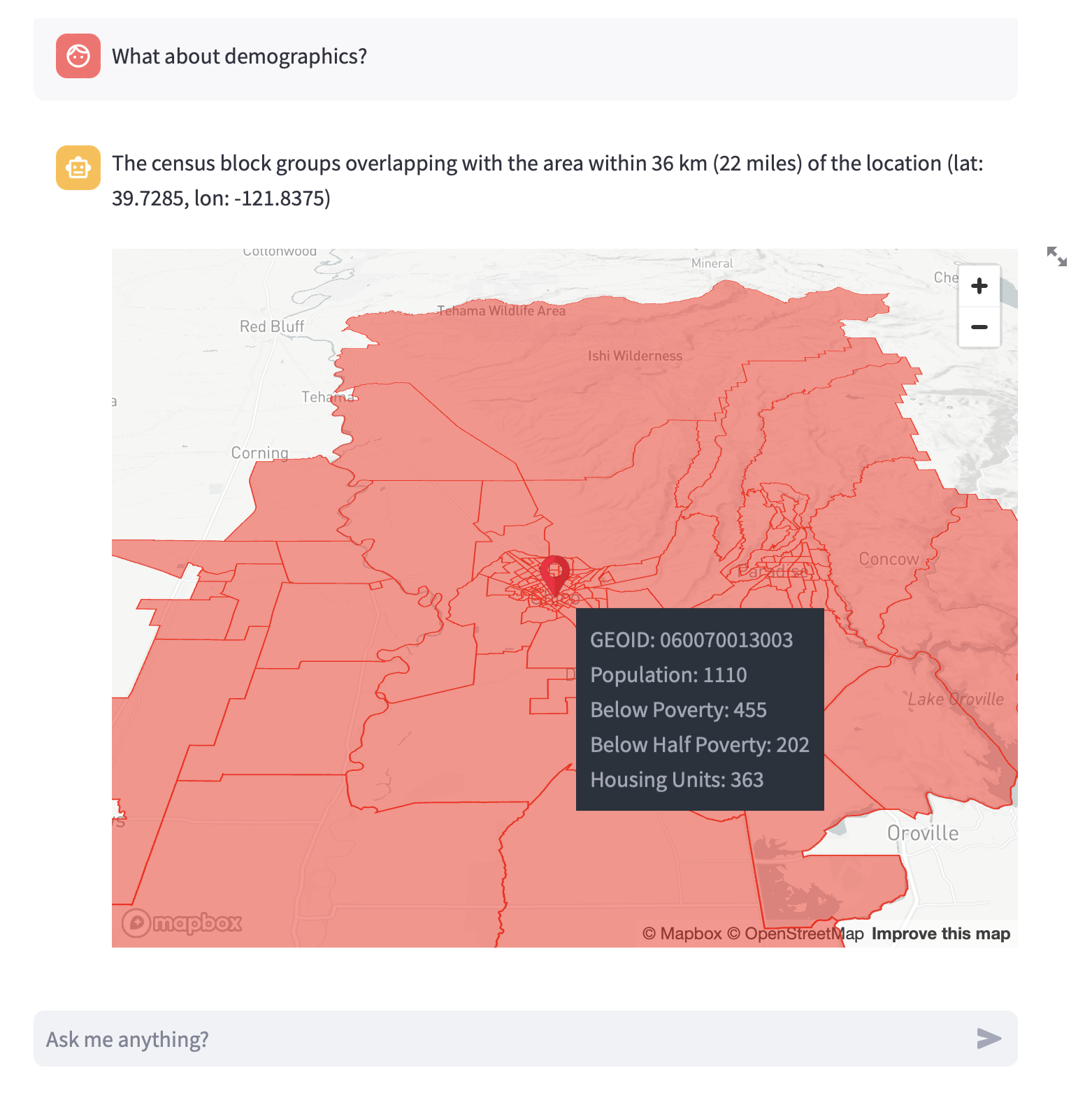}
    \caption{Interactive visualizations in the WildfireGPT user experience. This example is taken from the case study themed "Current Wildfire Risk Analysis." Users can select the season and time period to view the corresponding Fire Weather Index (FWI) map (left of first row),
    which displays risk levels using a color scale. Location-specific FWI values are accessible by hovering over the map (center of first row). By selecting different seasons and time period, the users can comprehend the changing landscape of the FWI (center of first row to center of second row). The wildfire incident map (right of the second row) shows the spatial distribution of recent fires, while the line graph (bottom left) presents the temporal trend of incidents. Socioeconomic data is visualized through census block group overlays (bottom right), providing insights into poverty rates and housing units in each area.}
\label{fig:wildfiregpt-interactive-vis}
\end{figure}

\subsection*{The WildfireGPT Experience}

\paragraph{User-Centered Multi-Agent Transition}

In our study we employ GPT-4 Turbo \citep{achiam2023gpt} as the backbone of WildfireGPT, enabling it to dynamically invoke different \edit{agents} via function calling through the OpenAI Assistant API \citep{openai2023assistant}. The user interacts with WildfireGPT through a chat interface built on the Streamlit-based web app \citep{streamlit}. \edit{Behind the scenes, we designed WildfireGPT as a multi-agent system where specialized agents collaborate under the coordination of a task orchestrator that routes interactions appropriately throughout the workflow. The integrated functions and multi-agent system architecture, including the implementation details of each specialized agent, are detailed in the Methodology section.}

From an end user perspective, WildfireGPT linearly transitions through three main \edit{agents}---the user profile \edit{agent}, the planning \edit{agent}, and the analyst \edit{agent}---to guide users through a seamless, interactive, and personalized experience (Figure \ref{fig:overview}). The user profile \edit{agent} gathers information by asking a series of predefined questions. These questions cover the user's professional background, primary concerns, location of interest, \edit{timeframe for} addressing concerns, and specific aspects of wildfire risks the user wants to explore, \edit{including desired outcomes}. Then, the \edit{agent} generates a detailed user profile and prompts the user to review and confirm the profile's accuracy. Based on the user profile, the planning \edit{agent} formulates a step-by-step action plan outlining the datasets to be analyzed, the focus of the literature review, and how the recommendations will be formulated \edit{by} the analyst \edit{agent}. The user can provide feedback on the plan, allowing for refinements until it meets the user's expectations, ensuring transparency and oversight. The analyst \edit{agent} guides the user through the analysis process outlined in the plan, presenting findings and recommendations, while actively addressing follow-up questions to ensure a comprehensive understanding and practical application of the results.

\paragraph{\edit{Diverse} Data Sources}
WildfireGPT combines \edit{natural hazard} projections, observational data, socioeconomic indicators, and scientific literature to deliver a multifaceted analysis of wildfire risks. Location-specific datasets include  FWI \cite{NWCG_FWI}) projections from the \edit{ClimRR portal} \citep{climrr_anl}, recent wildfire incident records (2015 to 2023) from the Wildland Fire Interagency Geospatial Services Group \citep{nifc2024wildlandfire, irwin2024wildfireinfo}, tree-ring and sediment-based fire history records from the International Multiproxy Paleofire Database \citep{margolis2021noaawds}, and census data on poverty rates and housing units from the 2022 American Community Survey 5-year estimates \citep{acs2022}. Furthermore, a vast collection of abstracts of scientific literature on wildfires, developed by Argonne National Laboratory for the Community and Infrastructure Adaptation to Climate Change tool, is also integrated \citep{mallick2023analyzing, mallick2024analyzing}.

\paragraph{Interactive Visualizations}

WildfireGPT provides interactive geospatial visualizations of the location-specific data within a 36 km radius of the user's area of interest, allowing the user to explore and analyze the data. For instance, the FWI displayed in Figure \ref{fig:wildfiregpt-interactive-vis} uses a color scale ranging from yellow (low risk) to red (very extreme). By hovering over a specific grid, the user can view the exact FWI value for that location; by selecting different seasons and time periods, the user can comprehend the changing landscape of the FWI. When visualizing recent wildfire history, the locations of all fire records are displayed as red dots on the map, and the temporal trend of wildfire incidents is shown in line graphs. Socioeconomic data, such as poverty rates and housing units, can be explored by hovering over census block groups. This comprehensive spatial visualization facilitates insights into the interplay among \edit{natural hazards}, wildfire risk, and socioeconomic factors, enabling informed decision-making and risk assessment.

\begin{table}[htb!]
\begin{small}
\centering
    \caption{Case studies conducted with domain experts to evaluate the performance and applicability of WildfireGPT across various professions, geographic locations, and wildfire management themes. \edit{Each case study represents either a location-specific inquiry derived from experts' ongoing research projects or a stakeholder perspective based on established regional collaborations. The description column outlines the specific focus of each case. } The studies span different time horizons and incorporate a range of data types and analytical approaches to assess wildfire risk, impact, and mitigation strategies. Case studies are listed in alphabetical order.\\}
    \label{tab:case_studies}
    \small
    \begin{tabular}{|p{0.12\linewidth}|p{0.1\linewidth}|p{0.1\linewidth}|p{0.58\linewidth}|}
    \hline
    \textbf{Theme} & \textbf{Profession} & \textbf{Location} & \textbf{Description} \\
    \hline
    Climate Impact on U.S. Wildfires & Atmospheric Scientist & Boston, MA & Exploring medium-term (10--30 years) impact of climate on wildfire occurrences and intensity in the U.S. Analyzing studies showing an increase in wildfire occurrences in Boston. Identifying and addressing uncertainties affecting future wildfire risks.\\
    \hline
    Comprehensive Wildfire Impact & Engineer & Mount Bigelow, AZ & Exploring long-term (50+ years) wildfire frequency, intensity, and impacts. Investigating mitigation strategies, ecological and economic effects, and infrastructure considerations.\\
    \hline
    Current Wildfire Risk Analysis & Data Analyst & Chico, CA & Analyzing current wildfire data, causes, and forest management practices. Exploring predictive models incorporating real-time variables. Gathering detailed demographic and socioeconomic profiles to assess risk and impact. \\
    \hline
    Ecosystem Fire Management & Landscape Manager & Naperville, IL & Analyzing wildfire frequency and impacts on vegetation with a focus on oak ecosystems. Developing strategies to balance fire use for oak ecosystem management with risk reduction for residential areas. \\
    \hline
    Hazard Mitigation Planning & Hazard Mitigation Planner & Mora County, NM & Assessing wildfire likelihood, frequency, intensity, and timing over the next 15 years for community hazard mitigation planning. \\
    \hline
    Infrastructure Wildfire Risk & Climate Change Risk Analyst & Denver, CO & Analyzing historical wildfire data and future projections to assess risks to energy sector infrastructure, considering land use changes and climate models (RCP 8.5 scenario). \\
    \hline
    Post-Wildfire Public Safety & Public Safety Manager & Sangre De Cristo Mountains, NM & Focusing on short-term (1--10 years) mitigation strategies after a high-intensity fire. Assessing emergency services readiness, community preparedness programs, and environmental restoration initiatives with the U.S. Army Corps of Engineers on erosion and flooding mitigations.\\
    \hline
    Private Property Protection & Homeowner & Near Covington, VA & Developing a 5--10-year plan for managing forest health, maximizing marketable species (focus on oak and cherry), and protecting properties from wildfires. \\
    \hline
    Urban Wildfire Mitigation & Urban Planner & Beaverton, OR & Developing a 30-year community plan focused on building codes and infrastructure resilience to mitigate fire-related hazard risks. \\
    \hline
    Wildland Urban Interface Impact & Risk Manager & Las Vegas, NM & Assessing medium-term (10--30 years) wildfire risks to housing and water resources in the wildland-urban interface. Exploring historical data, predictive models, current housing structures, future development plans, and wildfire impact on water quality and availability. Examining existing and proposed mitigation strategies.\\
    \hline
    \end{tabular}
\end{small}
\end{table}

\subsection*{Evaluating WildfireGPT: A Framework Built on Real-World Case Studies}

\edit{Evaluating conversational systems in the natural hazard domain presents unique challenges, including validating information from diverse datasets, handling diverse user queries across varying levels of expertise and backgrounds, assessing hazard conditions and their impact on different critical infrastructure, ensuring the accuracy and reliability of responses based on the user’s profile and background, and maintaining contextual relevance. While conversational systems are often evaluated by using lexical similarity metrics (e.g., BLEU \cite{papineni2002bleu}, ROUGE \cite{lin2004rouge}), they correlate poorly with human expert assessment of domain-specific content generation \cite{liu2021meta, deriu2022probing}. To address this limitation, we adopted an evaluation approach that prioritizes expert judgment over purely automated metrics. As part of this effort,} we conducted 10 in-depth case studies with domain experts \edit{from Argonne National Laboratory specializing in natural hazard resilience, wildfire risk management, and infrastructure vulnerability assessment. These domain experts participated in one-hour structured interviews, during which they directly interacted with WildfireGPT through a facilitated interface. The experts either posed questions from their ongoing projects or represented stakeholder perspectives based on their regional collaborations, providing real-time feedback on system performance. These case studies encompassed a diverse range of wildfire-related topics and locations across the United States, with details summarized in Table \ref{tab:case_studies}.}

\edit{Our evaluation of WildfireGPT employs three complementary stages. First, we conducted a modular comparison of the analyst agent's data and literature retrieval effectiveness against two alternatives—ChatClimate \cite{vaghefi2023chatclimate} and Perplexity AI \cite{perplexity2024}—using controlled simulations based on our case studies. Second, we performed a qualitative ablation study examining how different levels of user profile specificity impact the specificity and actionability of WildfireGPT's responses, also through controlled simulations derived from case study topics. Third, we evaluated the actual responses produced during expert case studies through: (1) expert assessment of response relevance, entailment, and accessibility and (2) exploration of scaling the human-in-the-loop evaluation using LLM-as-a-judge.\footnote{Our evaluation framework focuses on response quality and retrieval performance, as user intent identification is intrinsically validated through the \edit{user} profile agent and planning agent's confirmation protocol with the user---a design choice that differs from traditional conversational systems \cite{farshidi2024understanding}. Both profile and planning components incorporate explicit user verification of generated summaries before proceeding, embedding intent validation directly within the interaction pipeline.}}

\paragraph{Comparative Evaluation against Other Models}
\edit{
To evaluate the effectiveness of data retrieval and evidence-based query response capabilities, we compare WildfireGPT against two baseline alternatives practitioners might consider: ChatClimate and Perplexity AI. ChatClimate is a conversational AI platform developed to enhance understanding of natural hazard risks by providing accessible information grounded in the IPCC reports;  Perplexity AI is an AI-powered answer engine that delivers real-time responses by summarizing information from multiple web sources. Specifically, we extract 20 prompts from our case studies to test whether each method is able to (i) retrieve and interpret location-specific wildfire data and (ii) provide evidence-based answers to domain-specific questions. The evaluation metrics for data analysis include data provision (successful retrieval of relevant data), location specificity (geographical precision of retrieved data), and data accuracy (alignment between retrieved data and prompt requirements). Similarly, we compare the integration of external knowledge, the citation practices, the authority of the source, and the contextual relevance of responses to domain-specific questions.}

\paragraph{Personalization}
\edit{To investigate the effects of the user profile agent, we conduct a qualitative, two-phase ablation study to systematically evaluate the impact of user profile granularity and diversity on WildfireGPT's recommendations. Inspired by the Private Property Protection case study, for both phases we use the prompt ``Develop recommendations to enhance forest resilience against wildfires" to elicit a response from WildfireGPT. In the first phase, we systematically test three levels of user profile specificity provided to the agent: no profile information provided, location plus timeline, and the full profile of a power grid manager focused on maintaining transmission line clearance and grid resilience for power distribution reliability and access. In the second phase, we hold location and timeline constant and test five distinct professional profiles---homeowner, civil engineer, ecologist, emergency manager, and power grid manage---each with minimally distinct concerns and the respective scopes. Table \ref{tab:profiles} summarizes the differences in primary concerns and scopes across these professional profiles, with more detailed profile descriptions available in the Methodology section. We conduct a controlled simulation for each profile by passing the profile directly into WildfireGPT's planning and analyst agents, allowing us to observe how the same prompt generated tailored recommendations across different user types. Outputs are analyzed across three dimensions: \textit{plan generation}, \textit{literature retrieval}, and \textit{recommendation specificity}. 
}

\begin{table}[htbp]
\centering
\caption{User profile variations and literature search queries in Phase 2 of the WildfireGPT personalization ablation study. We hold location and timeline constant and test five distinct professional profiles, with changes in the concerns and the respective scopes. The search query represents the resulting literature retrieval query automatically generated by WildfireGPT's analyst agent.}
\begin{tabular}{|p{0.1\textwidth}|p{0.19\textwidth}|p{0.17\textwidth}|p{0.44\textwidth}|}
\hline
\textbf{Profession} & \textbf{Primary Concern} & \textbf{Scope} & \textbf{Search Query} \\
\hline
Homeowner & Maximizing marketable species & Health and marketable species & ``Strategies for managing forests to maintain health, maximize marketable species, and minimize wildfire risks in Virginia'' \\
\hline
Civil Engineer & Ensuring structural and infrastructural resilience & Drainage efficiency and slope stability & ``Wildfire risks and climate change impacts on forest management near Covington, VA; Strategies for enhancing drainage efficiency and slope stability; Structural resilience against wildfires in forested areas'' \\
\hline
Ecologist & Maintaining biodiversity and ecosystem services & Ecological resilience and habitat connectivity & ``Wildfire management and ecological resilience in forest ecosystems near Covington, VA'' \\
\hline
Emergency Manager & Establishing defendable space and evacuation corridors & Emergency access and response capabilities & ``Effective forest management practices, defendable space creation, evacuation protocols, and property protection measures against wildfires near Covington, VA'' \\
\hline
Power Grid Manager & Maintaining transmission line clearance and grid resilience & Power distribution reliability and access & ``Effective strategies for vegetation management, forest health maintenance, and wildfire risk mitigation around power grids near Covington, VA'' \\
\hline
\end{tabular}
\label{tab:profiles}
\end{table}

\begin{table}[htbp]
\centering
\caption{Domain expert evaluation criteria and corresponding questions used to assess WildfireGPT's performance in generating responses. Each question was rated on a three-point scale: "Yes" (1), "Could be better" (0.5), and "No" (0).\\}
\label{tab:human_eval_criteria}
\begin{tabular}{|l|l|}
\hline
\textbf{Criteria} & \textbf{Questions} \\
\hline
\multirow{6}{*}{Relevance} & (1) Does my response answer your last question? \\
 & (2) Is my response relevant to your profession? \\
 & (3) Is my response relevant to your concern? \\
 & (4) Is my response relevant to your location? \\
 & (5) Is my response relevant to your timeline? \\
 & (6) Is my response relevant to your scope? \\
\hline
Entailment & (1) Do my analyses or recommendations logically follow from the \\
 & information (data, literature) provided? \\
\hline
\multirow{3}{*}{Accessibility} & (1) Does my response contain too much jargon? \\
 & (2) Does my response provide enough explanation? \\
 & (3) Does my response contain redundant or useless information? \\
\hline
\end{tabular}
\end{table}

\paragraph{Domain Expert Evaluation}
During the case studies, whenever WildfireGPT generates a response, domain experts \edit{from Argonne National Laboratory's Environmental Science Division and Decision and Infrastructure Sciences Division} are asked to evaluate its relevance, entailment, and accessibility using a structured questionnaire. These experts are scientists and engineers specializing in \edit{natural hazards} and wildfire control with different detailed concentrations or other related professions as listed in Table~\ref{tab:case_studies}. Each question is rated on a three-point scale, with "Yes" receiving a score of 1, "Could be better" a score of 0.5, and "No" a score of 0. The valuation criteria and corresponding questions are presented in Table \ref{tab:human_eval_criteria}. Relevance assesses whether the model's responses appropriately address the user's last question and are relevant to the user's profession, concerns, location, timeline, and scope. Entailment evaluates the logical coherence between the model's analyses/recommendations and the provided data or literature. Accessibility examines the clarity and concision of the model's language, considering factors such as jargon, explanatory detail, and redundancy. We report the total scores for each criterion across all case studies. To complement the quantitative analysis, we also report qualitative insights from the expert evaluations to provide a more nuanced understanding of WildfireGPT's performance. 

\edit{We additionally verify whether the reported data and citations in WildfireGPT's responses originate from the retrieved sources. Complete verification results demonstrating WildfireGPT's consistently high information fidelity are provided in the Supplementary Materials.}

\paragraph{LLM-as-a-Judge Evaluation} To explore the potential for scalable, automated evaluation that could maintain quality assurance during deployment without constant human oversight, we implemented an LLM-as-a-judge approach. This method involves using a separate GPT-4 \cite{achiam2023gpt} query to assess the quality of WildfireGPT's responses based on the same criteria used in the expert evaluation. We report the agreement of the LLM-as-a-judge evaluation with the expert evaluation to determine the feasibility of using this approach. 
Each case study involved a unique interaction between the user and WildfireGPT, resulting in varying uses of data sources and different numbers of rounds of follow-up interactions. Our evaluation framework was designed to accommodate this variability, ensuring that the assessment remained meaningful and relevant to each specific case.

\begin{table}[htb!]
\begin{small}
\centering
\caption{Comparison of wildfire analysis tools for data analysis. 
WGPT = WildfireGPT, CC = ChatClimate, PAI = Perplexity AI. 
Data accuracy could not be evaluated for ChatClimate because there is no way to upload data files to its interface.
Each row represents a specific prompt or task related to wildfire analysis. For each metric, a $\checkmark$ indicates that the tool met the criteria for that prompt, while a $\times$ indicates it did not. NA means the metric was not applicable or could not be evaluated, typically because of the previous criteria not being met; for example, when data is not provided, it is not applicable to discuss the specificity or accuracy of the data. 
The superscript $^1$ indicates that although no specific number was mentioned, ChatClimate nonetheless interprets the increase in fire risks in words.}
\label{tab:comparison-data}
\resizebox{\textwidth}{!}{%
\begin{tabular}{|p{6cm}|ccc|ccc|cc|}
\hline
\multirow{2}{*}{\textbf{Prompt}} & \multicolumn{3}{c|}{\textbf{Data Provision}} & \multicolumn{3}{c|}{\textbf{Location Specificity}} & \multicolumn{2}{c|}{\textbf{Data Accuracy}} \\
\cline{2-9}
 & \textbf{WGPT} & \textbf{CC} & \textbf{PAI} & \textbf{WGPT} & \textbf{CC} & \textbf{PAI} & \textbf{WGPT} & \textbf{PAI} \\
\hline
(1) Analyze future climate data to understand potential trends in fire danger in Boston, Massachusetts & $\checkmark$ & $\checkmark^1$ & $\checkmark$ & $\checkmark$ & $\times$ & $\checkmark$ & $\checkmark$ & $\times$ \\
\hline
(2) Analyze the recent fire incident data for Mora County, New Mexico, reviewing wildfire frequency, intensity, and timing from 2015 to 2023 & $\checkmark$ & $\times$ & $\checkmark$ & $\checkmark$ & NA & $\checkmark$ & $\checkmark$ & $\times$ \\
\hline
(3) Review the long-term fire history records to assess the frequency and intensity of past wildfires over the last 50+ years in Mount Bigelow, Arizona & $\checkmark$ & $\times$ & $\times$ & $\checkmark$ & NA & NA & $\checkmark$ & $\checkmark$ \\
\hline
(4) Analyze demographic and socioeconomic profiles of the Chico area in California to understand which groups are most vulnerable in wildfire scenarios & $\checkmark$ & $\checkmark$ & $\checkmark$ & $\checkmark$ & $\times$ & $\checkmark$ & $\checkmark$ & $\checkmark$ \\
\hline
(5) Analyze long-term fire history records to assess wildfire events over the past 150 years in the Naperville, Illinois, region & $\times$ & $\times$ & $\times$ & NA & NA & NA & NA & NA \\
\hline
(6) Examine climate models and projections under the RCP 8.5 scenario to forecast potential future wildfire risks in the medium term (10--30 years) in Denver, Colorado & $\checkmark$ & $\checkmark$ & $\checkmark$ & $\checkmark$ & $\times$ & $\times$ & $\times$ & NA \\
\hline
(7) Analyze the recent fire incident data from the Wildland Fire Interagency Geospatial Services Group to assess the frequency, intensity, and locations of recent wildfires in Sangre de Cristo Mountains & $\checkmark$ & $\times$ & $\times$ & $\checkmark$ & NA & NA & $\checkmark$ & $\times$ \\
\hline
(8) Analyze the recent fire incident data from the last five years in Covington, Virginia & $\checkmark$ & $\times$ & $\checkmark$ & $\checkmark$ & $\times$ & $\checkmark$ & $\checkmark$ & $\times$ \\
\hline
(9) Analyze the FWI focusing on both current trends and projections for the mid-century period (2045--2054) to assess the potential increase in fire danger due to climate change in Beaverton, Oregon & $\checkmark$ & $\checkmark$ & $\checkmark$ & $\checkmark$ & $\times$ & $\times$ & $\checkmark$ & $\times$ \\
\hline
(10) Analyze projected wildfire risk data for the mid-century period (2045--2054) to understand the potential trends in fire danger in Las Vegas, New Mexico & $\checkmark$ & $\checkmark^1$ & $\checkmark$ & $\checkmark$ & $\times$ & $\times$ & $\checkmark$ & $\times$ \\
\hline
\textbf{Overall Success Rate} & \textbf{9/10} & \textbf{5/10} & \textbf{7/10} & \textbf{9/9} & \textbf{0/6} & \textbf{4/7} & \textbf{8/9} & \textbf{2/8} \\
\hline
\end{tabular}%
}
\end{small}
\end{table}

\subsection*{Comparative Evaluation against Baseline Models}

\edit{The comparative evaluation incorporated 20 prompts derived from the case studies: (i) location-specific data analysis tasks ($n=10$) requiring retrieval and interpretation of wildfire risk data (Table \ref{tab:comparison-data}); and (ii) evidence-based question-answering tasks ($n=10$) necessitating domain-specific knowledge retrieved from the corpus of scientific literature (Table \ref{tab:comparison-qa}). The corresponding case study of each prompt is detailed in the Supplementary Materials.}

\edit{The comparative data analysis in Table \ref{tab:comparison-data} reveals distinct performance patterns across the three models WildfireGPT, ChatClimate, and Perplexity AI in the categories of (1) data provision—successful retrieval of wildfire data; (2) location specificity—geographical precision of retrieved data; and (3) data accuracy—correctness of the retrieved data values when compared with reference datasets, evaluable only for WildfireGPT (with direct database access) and Perplexity AI (where external data files could be uploaded for analysis). In the data provision category, WildfireGPT significantly outperforms the others, succeeding in 9 out of 10 prompts, indicating strong capability in retrieving relevant wildfire-related data. Perplexity AI follows with 7/10, while ChatClimate trails with just 5/10. In terms of location specificity, WildfireGPT again leads with 9/9, demonstrating precise geographic referencing in its responses. Perplexity AI achieves moderate performance with 4/9, whereas ChatClimate fails entirely in this category (0/6). For data accuracy, WildfireGPT shows a clear advantage, with 8/9 correct outputs compared with Perplexity AI’s 2/8, indicating a substantial gap in the factual correctness of retrieved data. Note that while our initial evaluation framework included 10 test prompts, the effective sample size varies across categories because of dependencies between evaluation criteria. For instance, location specificity could  be assessed only when data provision was successful, and accuracy could be evaluated only when the  provided data contained specific locations. Overall, WildfireGPT consistently shows the highest performance across all three categories, making it the most robust tool for wildfire data analysis among those compared.}

\edit{WildfireGPT demonstrated superior performance in location-specific data retrieval due to the system's user profile agent, which systematically verifies exact geographical coordinates with users prior to data retrieval attempts. This feature ensures precise location identification, enabling successful data retrieval whenever the requested information exists within the database. The failure occurred in the Infrastructure Wildfire Risk case study (Denver, CO), where WildfireGPT provided Fire Weather Index analysis using RCP 4.5 rather than the requested RCP 8.5 projections because RCP 4.5 was the only scenario available in the dataset, although this limitation was not acknowledged when WildfireGPT interacted with the user—an issue that could be addressed in future iterations. 
ChatClimate exhibited significant analytical constraints, as its responses systematically defaulted to broad climatological generalizations without geographical precision, consistently failing to tailor information to the location in user queries. For example, in analyzing Boston's fire danger trends, ChatClimate inappropriately expanded its analytical scope to Arctic and Western U.S. fire patterns. This systematic geographical decontextualization significantly compromised the system's utility for location-specific wildfire risk assessment. This issue stems in part from ChatClimate’s primary reliance on IPCC reports and a limited repository of studies within its database, which prioritize global or regional-scale climate insights over localized, granular data.
Perplexity AI demonstrated reasonable performance in data provision but exhibited substantial limitations in location specificity when the corresponding data file was not provided by the user. The system primarily leverages publicly accessible information rather than conducting direct data analysis, resulting in significant geographical variance in performance based on public data availability. In regions with robust public documentation, such as Chico, CA, the system successfully integrated multiple data sources including local hazard mitigation plans. Conversely, analysis of areas with limited public records resulted in either null responses or inappropriate source application: for example, the system erroneously applied European wildfire danger reports to analyze local fire risk in Beaverton, OR. When the user uploads the data file, Perplexity AI incorrectly represents data extracted at geographical coordinates different from those specified in the query while claiming spatial correspondence or claims that the provided data does not include the requested information.}

\edit{The Ecosystem Fire Management case study (Naperville, IL) provides a particularly instructive methodological comparison when confronting identical data limitations. When prompted to analyze 150-year fire history records for Naperville, all three systems encountered data unavailability. WildfireGPT's response demonstrated adaptability by explicitly acknowledging the absence of historical data and proposing alternatives---specifically recommending the user to explore recent fire incident data and FWI projections to maintain analytical continuity despite the constraint. ChatClimate responded with generalization, discussing broad climatic shifts and anthropogenic interventions affecting wildfire patterns over the last 150 years without any Naperville-specific information, effectively abandoning the geographical aspects within the query. Perplexity AI acknowledged the data limitation but methodologically pivoted to institutional historiography, discussing the Naperville Fire Department's organizational history rather than wildfire patterns---a response that maintained geographical specificity while deviating from the analytical framework of the original query. These differential responses to identical data constraints reveal distinct epistemological approaches to knowledge gaps that systematically influence analytical utility.
}

\begin{table}[htb!]
\centering
\caption{Comparison of Wildfire analysis tools for evidence-based question qnswering. 
WGPT = WildfireGPT, CC = ChatClimate, PAI = Perplexity AI. 
Each row represents a specific prompt or task related to wildfire analysis. For each metric, a $\checkmark$ indicates that the tool met the criteria for that prompt, while a $\times$ indicates it did not. NA means the metric was not applicable or could not be evaluated, typically because of the previous criteria not being met; for example, when external knowledge is not retrieved, no citation would be provided. For ChatClimate, when it refers to studies outside of IPCC reports, it fails to provide the full title/link to the cited study.
Contextual relevance means that the response is related to the location to the best specificity possible and tries to address the prompt. 
The main reason behind ChatClimate's failures is that it often references general trends in large countries without a clear clue as to whether the information could be relevant/transferred to the location-specific queries in the prompt.
The superscript numbers indicate the following issues with PAI: 
$^1$Used articles from non-profit organizations, 
$^2$Included articles from news media, 
$^3$Used articles from commercial sources, 
$^4$Referenced a Reddit post.}
\label{tab:comparison-qa}
\begin{threeparttable}
\resizebox{\textwidth}{!}{%
\begin{tabular}{|p{5.5cm}|ccc|ccc|ccc|ccc|}
\hline
\multirow{2}{*}{\textbf{Prompt}} & \multicolumn{3}{c|}{\textbf{External Knowledge Retrieved}} & \multicolumn{3}{c|}{\textbf{Citation Provision}} & \multicolumn{3}{c|}{\textbf{Authoritative Sources}} & \multicolumn{3}{c|}{\textbf{Contextual Relevance}} \\
\cline{2-13}
 & \textbf{WGPT} & \textbf{CC} & \textbf{PAI} & \textbf{WGPT} & \textbf{CC} & \textbf{PAI} & \textbf{WGPT} & \textbf{CC} & \textbf{PAI} & \textbf{WGPT} & \textbf{CC} & \textbf{PAI} \\
\hline
Is FWI a robust estimate of the fire danger risk? & $\times$ & $\checkmark$ & $\checkmark$ & NA & $\checkmark$ & $\checkmark$ & NA & $\checkmark$ & $\checkmark$ & $\checkmark$ & $\checkmark$ & $\checkmark$ \\
\hline
What universities would make good partners to undertake studies on fire behavior and effective mitigation strategies in Mora County, NM? & $\times$ & $\times$ & $\checkmark$ & NA & NA & $\checkmark$ & NA & NA & $\checkmark$ & $\checkmark$ & $\checkmark$ & $\checkmark$ \\
\hline
What are the wildfire mitigation strategies, ecological impacts, and economic effects in forested areas similar to Mount Bigelow, AZ? & $\checkmark$ & $\checkmark$ & $\checkmark$ & $\checkmark$ & $\times$ & $\checkmark$ & $\checkmark$ & $\checkmark$ & $\times$\tnote{1} & $\checkmark$ & $\times$ & $\checkmark$ \\
\hline
Please provide sources on how to obtain data for Vegetation Type and Density, Previous Burn History, Human Activity Levels, Topography, Weather Conditions, Land Management Practices, Infrastructure and Accessibility, Socio-Economic Factors, and Climate Change Indicators & $\times$ & $\checkmark$ & $\checkmark$ & NA & $\checkmark$ & $\checkmark$ & NA & $\checkmark$ & $\checkmark$ & $\checkmark$ & $\checkmark$ & $\checkmark$ \\
\hline
It looks like March is the time folks implement a controlled burn in Naperville, IL. What do you think? & $\checkmark$ & $\times$ & $\checkmark$ & $\checkmark$ & NA & $\checkmark$ & $\checkmark$ & NA & $\checkmark$ & $\checkmark$ & $\times$ & $\checkmark$ \\
\hline
Let's dive deeper into Firebreaks and Buffer Zones. I'm interested in learning about approaches to establish firebreaks around the critical infrastructure and vegetation management practices. Can you provide more detail with evidences? & $\checkmark$ & $\checkmark$ & $\checkmark$ & $\checkmark$ & $\times$ & $\checkmark$ & $\checkmark$ & $\checkmark$ & $\checkmark$ & $\checkmark$ & $\times$ & $\checkmark$ \\
\hline
Search for scientific literature on effective post-wildfire recovery strategies, focusing on public safety and environmental restoration in Sangre De Cristo Mountains. & $\checkmark$ & $\checkmark$ & $\checkmark$ & $\checkmark$ & $\times$ & $\checkmark$ & $\checkmark$ & $\checkmark$ & $\times$\tnote{2} & $\checkmark$ & $\times$ & $\checkmark$ \\
\hline
Is fire common in this region prior to 1900s in Covington, VA? What is the natural fire regime? & $\checkmark$ & $\times$ & $\checkmark$ & $\checkmark$ & NA & $\checkmark$ & $\checkmark$ & NA & $\checkmark$ & $\checkmark$ & $\times$ & $\checkmark$ \\
\hline
Are there supporting literature for Buffer Zones that communities can implement to manage wildfire risk? & $\checkmark$ & $\checkmark$ & $\checkmark$ & $\checkmark$ & $\times$ & $\checkmark$ & $\checkmark$ & $\checkmark$ & $\times$\tnote{3} & $\checkmark$ & $\checkmark$ & $\checkmark$ \\
\hline
Are there good examples of similar cities to Las Vegas, NM that might experience similar shifts in wildfire risk? What mitigation strategies are those cities implementing? & $\checkmark$ & $\checkmark$ & $\checkmark$ & $\checkmark$ & $\checkmark$ & $\checkmark$ & $\checkmark$ & $\checkmark$ & $\times$\tnote{4} & $\checkmark$ & $\times$ & $\checkmark$ \\
\hline
\textbf{Overall Success Rate} & \textbf{7/10} & \textbf{7/10} & \textbf{10/10} & \textbf{7/7} & \textbf{3/7} & \textbf{10/10} & \textbf{7/7} & \textbf{7/7} & \textbf{6/10} & \textbf{10/10} & \textbf{4/10} & \textbf{10/10} \\
\hline
\end{tabular}%
}
\end{threeparttable}
\end{table}

\edit{
For evidence-based question answering, we employed four  metrics: (1) external knowledge integration—incorporation of relevant information beyond vanilla language generation; (2) citation practices—provision of complete, verifiable bibliographic information; (3) source authority—utilization of peer-reviewed scientific literature versus non-academic sources; and (4) contextual relevance—application of information to the specific geographical, ecological, and socioeconomic characteristics of query locations. Table \ref{tab:comparison-qa} shows distinct patterns in scientific literature integration and source validation. By design, all methods demonstrated strong integration of external knowledge (7/10 to 10/10). When external knowledge was integrated, both WildfireGPT and Perplexity AI provided complete citations of the sources, while ChatClimate exhibited bibliographic deficiencies (7/10 in citation provision). Despite using authoritative sources, ChatClimate uses only in-text parenthetical citations—author surname and year for scientific literature and standardized section references for IPCC documents (e.g., ``IPCC\_AR6\_WGII\_Chapter13, p.20")--without providing complete bibliographic lists. While readers familiar with climate science literature might recognize and locate IPCC citations, most scientific sources remain difficult to verify without full bibliographic information. WildfireGPT and ChatClimate relied on authoritative sources (7/7). Perplexity AI, on the other hand, demonstrated variable source authority (6/10), frequently incorporating non-peer-reviewed content including commercial publications, news media, non-profit organizations, and social media sources. }

\edit{Contextual relevance varied substantially across systems. WildfireGPT maintained high contextual relevance (10/10), providing literature and analyses that shared characteristics with the specified locations, whether through direct geographical proximity (same state or region) or ecological similarity (comparable forest types, fire regimes, or climate patterns). For example, in responses to queries about Sangre de Cristo Mountains and Mount Bigelow, WildfireGPT identified studies conducted in comparable montane forest ecosystems with similar elevation profiles and fire history patterns. Perplexity AI similarly achieved high contextual relevance (10/10), although with variable levels of authority in source selection. The contextual relevance can be attributed to both WildfireGPT and Perplexity AI integrating a large set of sources to allow rich information to be retrieved, although we identify areas for improvement in the following sections. In contrast, ChatClimate demonstrated substantial contextual relevance limitations (4/10), frequently defaulting to continental or hemispheric climate trends without establishing clear applicability to the specified locations. This geographical generalization was particularly evident in responses to queries about Las Vegas, NM, and Covington, VA, where the system referenced broad North American climate patterns without demonstrating specific relevance to the distinctive fire regimes of the southwestern high desert or Appalachian forest ecosystems, respectively.}

\edit{Overall, WildfireGPT demonstrated precise geographical analysis with methodological adaptation when confronting data limitations. ChatClimate exhibited geographical imprecision. Perplexity AI demonstrated comprehensive information retrieval with significant limitations in source validation and data interpretation accuracy. These performance patterns highlight fundamental distinctions in the analytical architecture of the three models. Furthermore, they demonstrate that having a wide but curated range of valid scientific resources with efficient retrieval mechanisms is critical for location-specific wildfire risk assessment and evidence-based domain-specific inquiry.}

\subsection*{Personalization}

\edit{Our two-phase ablation study explores the impact of user profile granularity and diversity on WildfireGPT's recommendations. In the first phase, we progressively test three levels of user profile specificity provided to the agent: no profile information provided, location plus timeline, and the full profile of a power grid manager focused on maintaining transmission line clearance and grid resilience for power distribution reliability and access. In the second phase, we hold location and timeline constant and test five distinct professional profiles---homeowner, civil engineer, ecologist, emergency manager, and power grid manage---each with minimally distinct concerns and the respective scopes. We report the results from both phases below.}

\noindent\textbf{Hierarchical Specificity from Profile Granularity} 
\edit{Table \ref{tab:phase one} demonstrates the tiered progression in recommendation specificity provided by WildfireGPT in the first phase of our ablation study. The table focuses on three key aspects of wildfire resilience---vegetation management, technology implementation, and role-specific action---and demonstrates how recommendations become increasingly targeted and context-aware as input specificity increases. For example, a generic prompt without profile information yields broad guidance such as thinning and pruning or the use of satellite imaging and drones. When location and timeline details are included, the system generates more ecologically informed and time-sensitive strategies, such as "develop and promote fire-adapted ecosystems by incorporating native species." With a complete power grid manager profile, the recommendations become more specific and operational, with recommendations prioritizing "sensors and advanced monitoring systems along the grid" and "expand the Right-of-Way (ROW) clearance around transmission lines," synthesizing ecological approaches with sector-specific operational requirements. Most notably, technological recommendations progressed from general "satellite imaging and drones" to grid-specific applications for "real-time surveillance of power lines, especially in hard-to-reach areas." More aspects are reported in the Supplementary Materials. This stratification demonstrates how increasing profile granularity introduced by the user profile agent enables WildfireGPT to align general resilience principles with the operational needs of distinct professional roles, particularly in critical infrastructure contexts where vegetation management intersects with operational reliability requirements.}

\begin{table}[htb!]
\centering
\caption{Comparative analysis across three profile specificity tiers. WildfireGPT responses demonstrate progressive domain adaptation from generic principles to stakeholder-specific protocols when provided with increasing user profile granularity. Representative outputs shown for vegetation management, technology implementation, and role-specific interventions demonstrate systematic enhancement of recommendation precision with profile enrichment. Italicized text highlights power-grid-specific recommendations emergent only with comprehensive user profiling. The full table can be found in the Supplementary Materials.}
\small
\begin{tabular}{|p{0.12\textwidth}|p{0.25\textwidth}|p{0.25\textwidth}|p{0.31\textwidth}|}
\hline
\textbf{Aspect} & \textbf{No Profile Information} & \textbf{+ Location and Timeline} & \textbf{+ Power-Grid Manager Profile} \\
\hline
\textbf{Vegetation Management} & 
``Thinning and Pruning: Reduce forest density through thinning and pruning [...] Controlled Burns: Implement controlled burns to reduce available fuels.'' & 
``Develop and promote fire-adapted ecosystems by incorporating native species [...] Practice selective cutting to reduce fire risk.'' & 
``Establish routine schedules for pruning and removing trees that {\it pose a risk to power lines}. Utilize drones or satellite imagery to monitor vegetation growth for timely intervention.'' \\
\hline
\textbf{Technology} & 
``Employ technologies like satellite imaging and drones [...] Use artificial intelligence to predict fire patterns.'' & 
``Invest in enhanced surveillance systems [...] satellite imagery, drones, and remote sensors to detect early signs of fire.'' & 
``Utilize Geographic Information Systems (GIS) and remote sensing [...] Install sensors and advanced monitoring systems along the grid to detect temperature changes, smoke, or increased humidity levels indicating the risk of fire. Use drones for real-time surveillance [...] of {\it power lines, especially in hard-to-reach areas.}'' \\
\hline
\textbf{Role-Specific Actions} & 
``International cooperation [...] Reduce carbon footprints.'' & 
``Encourage Responsible Logging Practices [...] Climate Adaptation Strategies.'' & 
``{\it Expand the Right-of-Way (ROW) clearance around transmission lines} beyond standard regulations [...] Implement an IVM program that combines mechanical, biological, and chemical methods to manage vegetation'' \\

\hline
\end{tabular}
\label{tab:phase one}
\end{table}

\begin{table}[htb!]
\centering
\caption{Distribution of Recommendation Categories Across Five Stakeholder Profiles. We use checkmarks to indicate which recommendations appear in each stakeholder’s profile.}
\begin{tabular}{|p{0.35\linewidth}|p{0.1\linewidth}|p{0.1\linewidth}|p{0.1\linewidth}|p{0.1\linewidth}|p{0.1\linewidth}|}
\hline
\textbf{Recommendation Category} & \textbf{Emergency Manager} & \textbf{Ecologist} & \textbf{Civil Engineer} & \textbf{Homeowner} & \textbf{Power Grid Manager} \\
\hline
\multicolumn{6}{|l|}{\textit{Fuel Management}} \\
\hline
Regular thinning/pruning/fuel reduction & \checkmark & \checkmark & \checkmark & \checkmark & \checkmark \\
\hline
Controlled/prescribed burns & \checkmark & \checkmark & \checkmark & \checkmark & \checkmark \\
\hline
\multicolumn{6}{|l|}{\textit{Infrastructure and Planning}} \\
\hline
Defendable space creation & \checkmark & & & \checkmark & \checkmark \\
\hline
Fire-resistant materials/infrastructure & \checkmark & \checkmark & & \checkmark & \checkmark \\
\hline
Emergency access/water sources & \checkmark & & \checkmark & \checkmark & \checkmark \\
\hline
\multicolumn{6}{|l|}{\textit{Community Engagement}} \\
\hline
Evacuation planning & \checkmark & \checkmark & & \checkmark & \\
\hline
Public education/awareness programs & \checkmark & \checkmark & \checkmark & \checkmark & \checkmark \\
\hline
\multicolumn{6}{|l|}{\textit{Collaboration and Management}} \\
\hline
Cross-boundary coordination & \checkmark & \checkmark & & & \checkmark \\
\hline
Monitoring systems/technology & \checkmark & \checkmark & \checkmark & \checkmark & \checkmark \\
\hline
Research & \checkmark & \checkmark &  &  & \checkmark \\
\hline
\multicolumn{6}{|l|}{\textit{Ecological Considerations}} \\
\hline
Fire-resistant species promotion & & & \checkmark & \checkmark & \\
\hline
Natural fire regime maintenance & & \checkmark & & & \\
\hline
Ecological monitoring & & \checkmark & \checkmark & & \checkmark \\
\hline
\end{tabular}
\label{tab:recommendations-comparison}
\end{table}

\noindent\textbf{Professional Identity as a Recommendation Filter }
\edit{
For the second phase, we summarize WildfireGPT's adaptation to different professional roles and priorities across three dimensions: \textit{plan generation}, \textit{literature retrieval} and \textit{recommendation specificity}. The full experimental results are included in the Supplementary Materias. }

\edit{First, our analysis shows that the planning agent adapts to the user's professional role and concerns. 
For example, when a power grid manager expresses primary concern about maintaining transmission lines and grid resilience, WildfireGPT proposes to search for literature about vegetation management and wildfire protection strategies around power grids. Similarly, when a homeowner indicates a focus on marketable species, WildfireGPT targets research about economically valuable tree species and property protection methods. This pattern continues across all profiles---emergency managers receive evacuation-focused plans, civil engineers get infrastructure-centered recommendations, and ecologists receive biodiversity-oriented strategies. In each case WildfireGPT maintains the same basic framework (analyzing fire data, reviewing literature, developing recommendations) but tailors the specific content to match the user's professional needs and objectives. This alignment between user profiles and generated plans is by design, as the planning agent's primary function is to establish clear expectations and a structured pathway through which the analyst agent can deliver personalized insights relevant to each stakeholder's specific context.}

\edit{Second, the analyst agent uses search queries calibrated to professional priorities when conducting literature retrieval, with details provided in Table \ref{tab:profiles}. For each profession, the search queries effectively capture their unique concerns and scope—homeowners' focus on marketable species, civil engineers' emphasis on structural resilience, ecologists' focus on ecological resilience, emergency managers' priority on defendable space, and power grid managers' concentration on grid resilience. These search queries distinctly different literature being retrieved, and we report the literature retrieval results based on these queries. When a power grid manager's profile emphasized transmission line clearance and grid resilience, the system yielded studies such as Hvenegaard's (2014) \cite{hvenegaard2014long} analysis of fuel treatment monitoring and Dale's (2006) \cite{dale2006wildfire} critique of fire suppression policies. Conversely, when responding to a civil engineer's profile, the system retrieved Stevens et al.'s (2020) \cite{stevens2020forest} findings on vegetation structure and soil moisture dynamics. This pattern of query-response alignment extends across all professional identities: emergency management queries yielded Ager et al.'s (2019) \cite{ager2019wildfire} work on community wildfire exposure, while the ecologist’s profile resulted in Schoennagel et al.'s (2017) \cite{schoennagel2017adapt} research on adaptive resilience and Waltz et al.'s (2014) \cite{waltz2014effectiveness} findings on ecological impacts of fuel reduction treatments. }

\edit{This adaptation pattern extends to recommendation development. As shown in Table \ref{tab:recommendations-comparison}, WildfireGPT generates distinct recommendation sets across the five professional profiles while maintaining core wildfire management principles. The table uses checkmarks to indicate which recommendations appear in each stakeholder's profile—for example, all profiles receive recommendations for regular thinning and prescribed burns, while only the ecologist receives natural fire regime maintenance recommendations. Although certain fundamental strategies—such as regular thinning/pruning and controlled burns—are universally recommended across all profiles, significant variations emerge in other categories based on professional priorities. For civil engineers, recommendations pivot toward drainage efficiency and slope stability considerations, incorporating Stevens et al.'s (2020) \cite{stevens2020forest} findings on soil moisture dynamics. 
Notably, civil engineers are the only profession not recommended to implement fire-resistant materials/infrastructure, instead focusing on ecological monitoring and emergency access solutions. Emergency managers receive the most comprehensive recommendation set, with checks in nearly all categories and exclusive emphasis on defendable space creation and evacuation planning. Ecologists uniquely receive recommendations for natural fire regime maintenance, reflecting their biodiversity conservation priorities, while being the only group besides civil engineers not advised on defendable space creation. Emergency management profiles elicit recommendations heavily weighted toward evacuation planning and defendable space creation, drawing on Ager et al.'s (2019) \cite{ager2019wildfire} community exposure analyses. Homeowners and power grid managers show similar recommendation patterns in infrastructure planning but diverge significantly in community engagement—homeowners receive evacuation planning advice while power grid managers do not. Interestingly, power grid managers share the research recommendation with emergency managers and ecologists, highlighting the technical expertise common to these roles.}

\edit{This pattern of differential adaptation suggests that user profiles serve as effective filters for distilling broadly applicable wildfire management principles into professionally actionable, context-specific recommendations. The profile-specific recommendations reflect the granularity in both tactical precision and implementation feasibility, aligned with each stakeholder's operational priorities while maintaining scientific rigor in the underlying management principles.}

\edit{Though the user profile agent help successfully retrieve thematically-relevant literature at a profession-specific level, the specificity of matches invites further refinement that goes beyond the user profile agent. First, when matching papers to user concerns, it might not align exactly with the specific scope. For example, when the power grid manager requested information about protecting power lines, the system identified Hvenegaard's (2014) \cite{hvenegaard2014long} paper on fuel treatment. This is valuable for fire management, but does not specifically address power grid infrastructure. Second, regarding location-specificity, searches for Covington, Virginia yielded Schoennagel et al.'s work in western North America or Hansen's research in Grand Teton National Park \citep{hansen2020can}—which could still offer applicable insights despite the geographic mismatch. These patterns reflect two areas for enhancement for the analyst agents: (i) the finite scope of the available literature database where closely aligned research may not exist for highly specialized user needs (such as power grid protection in specific geographic contexts); and (ii) inherent limitations in literature search mechanism as a part of the analyst agent. We explore these limitations and potential solutions in greater detail in the Discussion section.}

\subsection*{Domain Expert Evaluations: Relevance, Entailment, and Accessibility of WildfireGPT's Responses}

\begin{table}[ht]
\centering
\caption{Relevance scores for each case study, representing the percentage of responses that appropriately addressed the user's context and needs. Raw scores are shown (score/total times the domain expert opinion was collected to evaluate a response).\\ }
\label{tab:relevance_scores}
\begin{tabular}{|p{0.3\linewidth}|p{0.09\linewidth}p{0.09\linewidth}p{0.09\linewidth}p{0.09\linewidth}p{0.09\linewidth}p{0.09\linewidth}|}
\hline
\textbf{Case Study} & \textbf{Last Question} & \textbf{Profession} & \textbf{Concern} & \textbf{Location} & \textbf{Time} & \textbf{Scope} \\
\hline
Climate Impact on U.S. Wildfires & 17/18 & 18/18 & 18/18 & 18/18 & 18/18 & 18/18 \\
\hline
Comprehensive Wildfire Impact & 6/6 & 7/7 & 7/7 & 7/7 & 7/7 & 7/7 \\
\hline
Current Wildfire Risk Analysis & 5/5 & 6/6 & 6/6 & 6/6 & 5.5/6 & 6/6 \\
\hline
Ecosystem Fire Management & 3/3 & 2.5/3 & 3/3 & 3/3 & 3/3 & 3/3 \\
\hline
Hazard Mitigation Planning & 5/5 & 7/7 & 7/7 & 7/7 & 7/7 & 7/7 \\
\hline
Infrastructure Wildfire Risk & 7/7 & 7/7 & 7/7 & 7/7 & 7/7 & 7/7 \\
\hline
Post-Wildfire Public Safety & 3/3 & 4/4 & 4/4 & 3.5/4 & 4/4 & 4/4 \\
\hline
Private Property Protection & 6.5/7 & 8/9 & 9/9 & 8/9 & 9/9 & 9/9 \\
\hline
Urban Wildfire Mitigation & 3.5/4 & 4/4 & 4/4 & 4/4 & 4/4 & 4/4 \\
\hline
Wildland Urban Interface Impact & 2/2 & 4/4 & 4/4 & 4/4 & 3.5/4 & 4/4 \\
\hline
\textbf{Average across All Percentages} & 97.48\% & 97.22\% & 100.00\% & 97.64\% & 97.92\% & 100.00\% \\
\hline
\textbf{Overall Success Rate} & 96.67\% (58/60) & 97.83\% (67.5/69) & 100.00\% (69/69) & 97.83\% (67.5/69) & 98.55\% (68/69) & 100.00\% (69/69) \\
\hline
\end{tabular}
\end{table}

\begin{table}[ht]
\centering
\caption{Entailment and accessibility scores for each case study. Entailment indicates the percentage of responses where the model's analyses and recommendations logically followed from the provided information (data, literature). Accessibility assesses the clarity and concision of the model's language. Scores represent the percentage of responses that avoided jargon, provided sufficient explanation, and minimized redundancy. Raw scores are shown (score/total times the domain expert opinion was collected to evaluate a response).}
\label{tab:entailment_scores}
\begin{tabular}{|l|c|ccc|}
\hline
\multirow{2}{*}{\textbf{Case Study}} & \multirow{2}{*}{\textbf{Entailment}} & \multicolumn{3}{c|}{\textbf{Accessibility}} \\
\cline{3-5}
 &  & \textbf{No Jargon} & \textbf{Enough Explanation} & \textbf{No Redundancy} \\
\hline
Climate Impact on U.S. Wildfires & 11/13 & 17/17 & 16/17 & 14.5/17 \\
\hline
Comprehensive Wildfire Impact & 7/7 & 7/7 & 7/7 & 7/7 \\
\hline
Current Wildfire Risk Analysis & 5.5/6 & 6/6 & 5.5/6 & 6/6 \\
\hline
Ecosystem Fire Management & 3/3 & 3/3 & 3/3 & 3/3 \\
\hline
Hazard Mitigation Planning & 6/6 & 7/7 & 7/7 & 7/7 \\
\hline
Infrastructure Wildfire Risk & 6.5/7 & 7/7 & 7/7 & 7/7 \\
\hline
Post-Wildfire Public Safety & 3/3 & 4/4 & 4/4 & 4/4 \\
\hline
Private Property Protection & 6/6 & 8/8 & 7/8 & 7/8 \\
\hline
Urban Wildfire Mitigation & 1.5/2 & 4.5/5 & 2.5/5 & 5/5 \\
\hline
Wildland-Urban Interface Impact & 2.5/3 & 4/4 & 4/4 & 4/4 \\
\hline
\textbf{Average Across All Percentages} & 92.75\% & 99.00\% & 92.33\% & 97.28\% \\
\hline
\textbf{Overall Success Rate} & 92.86\%  & 99.26\% & 92.65\%  & 94.85\% \\
 &  (52/56) &  (67.5/68) &  (63/68) &  (64.5/68) \\
\hline
\end{tabular}
\end{table}

{\bf Relevance:} \edit{Domain experts generally found WildfireGPT's responses contextually appropriate (}a success rate of 96.67\% to 100\% across all relevance categories, Table \ref{tab:relevance_scores}), but there are areas for improvement. For the "Last Question" and "Profession" categories (Table \ref{tab:human_eval_criteria}, Q1 \& Q2), domain experts noted occasional misalignments, such as offering broad advice when specific zoning changes are requested or making impractical recommendations such as suggesting that homeowners host community workshops. The "Location" category revealed potential limitations in the literature corpus (Table \ref{tab:human_eval_criteria}, Q4), such as offering recommendations for homeowners in Virginia based on studies from the West Coast. WildfireGPT sometimes  failed to specify which strategies could be reasonably implemented within given timeframes (Table \ref{tab:human_eval_criteria}, Q5).

{\bf Entailment:} With an overall success rate of 92.86\% (Table \ref{tab:entailment_scores}), WildfireGPT generally remained logically consistent; but experts noted some issues when WildfireGPT provided recommendations, such as a lack of transparency in tracing the sources of its recommendations and a failure to present recommendations in a logical hierarchy from most important to least important. These situations could lead to long-winded responses that lack focus.

{\bf Accessibility:} WildfireGPT generally performed well in avoiding jargon (99.26\% overall success rate) and redundancy (92.65\% overall success rate), with most responses providing sufficient explanation (94.85\% overall success rate); see Table \ref{tab:entailment_scores}. However, the interpretation of these scores varied by audience. Urban planners and homeowners frequently requested more detailed explanations and examples as well as clearer definitions of technical terms and contextual information for data interpretation. In contrast, climate scientists preferred more concise responses, with less detail and more direct answers to their questions. These findings point to the challenge of balancing detail and concision to meet diverse user needs.

Overall, the domain expert evaluation demonstrates that WildfireGPT is capable of generating relevant, logically sound, and accessible responses to user queries across a range of wildfire-related scenarios. However, the results also highlight potential areas for improvement, such as improving the relevance of responses to the user's last question in certain cases and ensuring  differing levels of explanation to maximize clarity and utility for different domain experts.

\paragraph*{Domain Expert Evaluations: Qualitative Feedback on WildfireGPT's Recommendations}

To better understand the nuance and practical value of WildfireGPT's recommendations, we sought qualitative feedback from domain experts as they evaluated the system's responses via the structured questionnaire reported above. 

One notable example of WildfireGPT's effectiveness in integrating data analysis with domain-specific knowledge is demonstrated in the Ecosystem Fire Management case study (Figure \ref{box:Ecosystem_Fire_Management}). When asked for suggestions on adapting fire management strategies to maintain oak ecosystems while minimizing risks to residential areas, WildfireGPT analyzed recent wildfire data, revealing that controlled burns typically occur in March. It validated this response by citing literature such as that of Stan, Rigg, and Jones (2006) \citep{stan2006dynamics}, which emphasizes the importance of March fires for white oak regeneration. WildfireGPT also analyzed seasonal FWI data, identifying lower-risk periods in early spring and late autumn. Synthesizing these insights from more than one data source, it recommended timing controlled burns to coincide with historically low FWI values, balancing ecological management goals with safety. The domain expert praised WildfireGPT's nuanced approach, emphasizing its ability to balance ecological management goals with safety considerations.

\begin{figure}[hbt!]
\centering
\begin{tcolorbox}[title=Case Study: Ecosystem Fire Management]
\textbf{Context:} WildfireGPT analyzed recent wildfire incident data and found that from 2018 to 2021, the months with recorded incidents were March (2 incidents) and July (1 incident), suggesting that March may be when controlled burns are typically implemented. To confirm this point, WildfireGPT retrieved a study by Stan, Rigg, and Jones (2006) \citep{stan2006dynamics}, which highlighted the importance of reintroducing fire in March for white oak regeneration in northeastern Illinois. Additionally, WildfireGPT analyzed seasonal FWI data, identifying lower-risk periods during spring.\\
\textbf{User Query:} Please suggest specific adaptations to fire management strategies (to balance the utilization of fire for managing oak ecosystems and reducing the risk to residential and human developments.)

\textbf{WildfireGPT Response (Snapshot):}\\
\textit{Optimized Timing for Controlled Burns}:
\begin{itemize}
    \item Continue to utilize controlled burns in March when the FWI (Fire Weather Index) historically shows lower values, minimizing the risk of uncontrollable wildland fires. 
    \item Schedule additional controlled burns during early spring or late autumn to take advantage of similarly low FWI scores, ensuring burns are conducted under optimal conditions for controlling spread and smoke dispersal.
\end{itemize}

\textbf{Expert Comment:} The answer effectively addresses a key concern: whether to continue conducting controlled burns in March. It confirms that March remains a suitable time for burns due to historically lower Fire Weather Index (FWI) values, which reduce the risk of uncontrolled fires.
The response also provides a nuanced suggestion by recommending additional burn periods in early spring and late autumn. This approach takes advantage of similarly low FWI scores during these times, allowing for burns at both the beginning and end of the dormant season.
This strategy demonstrates a nuanced understanding of fire management, as it balances the need for regular burns with the importance of conducting them under optimal weather conditions for safety and control. The suggestion to utilize multiple burn windows throughout the year shows a thoughtful approach to ecosystem management while minimizing risks to nearby developments.
\end{tcolorbox}
\caption{Ecosystem Fire Management: WildfireGPT demonstrates its ability to integrate data analysis and domain knowledge to provide actionable recommendations for ecosystem fire management. By analyzing wildfire incident data, retrieving relevant literature, and examining seasonal FWI trends, WildfireGPT suggests optimizing the timing of controlled burns to minimize risks while maintaining oak ecosystem health. The domain expert's positive feedback highlights WildfireGPT's nuanced approach and its potential to support informed decision-making in wildfire management.}
\label{box:Ecosystem_Fire_Management}
\end{figure}

\begin{figure}[hbt!]
\begin{tcolorbox}[title=Case Study: Wildland-Urban Interface Impact]
\textbf{Context:}
In Las Vegas, NM, a risk manager developing urban plans for 2050 is concerned about the increasing wildfire risks and their impact on water resources. Previous analysis of Fire Weather Index (FWI) data indicated an increasing trend in fire danger levels, particularly during the spring season. \\
\textbf{User Query:} Proceed with the recommendations (on urban planning strategies to minimize wildfire risks and methods to protect water resources from the adverse effects of wildfires)\\
\textbf{WildfireGPT Response (Snapshot):}\\
\textit{Water Resource Management}
\begin{itemize}
    \item Water Supply Resilience: Increase the resilience of water supply systems to ensure availability during and after wildfire events. This could involve constructing new reservoirs, enhancing existing water treatment facilities, and establishing emergency water supply locations.
    \item Protect Water Quality: Develop strategies to protect water quality from the effects of wildfires, such as increased sedimentation and contamination. These might include erosion control measures and enhanced filtration systems at water treatment facilities. 
\end{itemize}

\textbf{Expert Comment:} The recommendation to protect water quality from increased sedimentation and contamination is highly relevant. This issue was a central concern when we initially developed this case study, and it remains a significant challenge for the area. The local water treatment plant is struggling to handle the fine-grained sediment that is now flowing downstream in much larger quantities than before. The plant's processes are overwhelmed by the volume of material, which far exceeds what they typically dealt with from normal erosion.
It's very good that the answer highlights the need to address water quality issues, specifically mentioning sedimentation and contamination. The suggestion to implement erosion control measures and enhance filtration systems directly addresses the problems we encountered. In our modeling work, we aimed to predict the sediment load to help design appropriate mitigation measures. The recommendation aligns well with the actual challenges faced in the field and the solutions we were exploring.

\textbf{User Query:} Are there good examples of similar cities that might experience similar shifts in wildfire risk? What mitigation strategies are those cities implementing?\\
\textbf{WildfireGPT Response (Snapshot):}\\
Here's a look at some of the approaches highlighted in recent research:[...] California is a prime example of a region adapting to increased wildfire risks. The study by Moritz and Stephens (2008) emphasizes the need for a sustainable coexistence with wildfire through policy, planning, and management changes. Key strategies include the following:
\begin{itemize}
    \item Risk-Based Framework: Implementing a risk-based framework for fire-related decisions to address natural hazards systematically.
    \item Reintroduction of Fire: Using controlled burns to reintroduce fire in a managed way to fire-prone ecosystems.
    \item Urban Planning Reevaluation: Reevaluating urban planning and building in fire-prone locations to make future cities less susceptible to wildfire damage.
\end{itemize}
\textit{References}:\\
Moritz, M., \& Stephens, S. (2008). Fire and sustainability: considerations for California's altered future climate. https://doi.org/10.1007/s10584-007-9361-1\\
\textbf{Expert Comment:} The WildfireGPT response effectively identifies California as a region experiencing similar shifts in wildfire risk to Las Vegas, NM. Despite some differences in fire characteristics, California remains a relevant comparison. The strategies highlighted, such as implementing a risk-based framework, reintroducing controlled burns, and reevaluating urban planning, could provide valuable guidance for Las Vegas.
\end{tcolorbox}
\caption{Wildland Urban Interface Impact: WildfireGPT demonstrates its effectiveness in addressing the complex challenges posed by the wildland-urban interface. Wildfire's recommendations regarding water quality protection align well with the actual challenges faced by domain experts in the field. WildfireGPT's ability to identify relevant case studies from other areas with similar shifts in wildfire risk and to highlight pertinent mitigation strategies showcases its potential to support informed decision-making in wildfire risk management and urban planning.}
\label{box:Wildland Urban Interface Impact on Housing and Water Resources}
\end{figure}

The Wildland Urban Interface Impact case study (Figure \ref{box:Wildland Urban Interface Impact on Housing and Water Resources}) further demonstrates WildfireGPT's practical utility. When the domain expert asked questions from the perspective of a risk manager from Las Vegas, NM, WildfireGPT provided recommendations on water resource protection that aligns well with actual challenges faced in the field. Furthermore, it identifies California as a relevant comparison due to similar fire characteristics. The retrieved study by Moritz and Stephens (2008) \citep{moritz2008fire} recommends strategies such as risk-based frameworks, controlled burns, and reevaluating urban planning---approaches the domain expert recognizes as potentially valuable for Las Vegas, NM. 

Overall, the qualitative feedback from domain experts underscores WildfireGPT's ability to provide nuanced recommendations that align with real-world challenges and existing research. We also provide more examples of qualitative feedback in the Supplementary Materials.

\paragraph{Scaling Up Evaluation by LLM-as-a-Judge}
To evaluate GPT-4's \cite{achiam2023gpt, liu2023g} feasibility as an automated judge for WildfireGPT, we compared GPT-4 assessments with human judgments on relevance, entailment, and accessibility. We applied LLM-as-a-judge to evaluate WildfireGPT’s responses only immediately after data or literature retrieval to manage context length and maintain performance. Agreement rates varied: relevance (62.99\%), entailment (75.00\%), and accessibility (66.67\%). Disagreements primarily occurred when experts rated responses positively ("Yes"), whereas GPT-4 suggested improvements ("Could be Better"), reflecting GPT-4’s stricter evaluative criteria.

\begin{table}[hbt!]
\centering
\caption{Agreement between human evaluation and LLM-as-a-Judge evaluation of WildfireGPT responses. Most misalignments occur when domain experts respond with "Yes" while GPT-4 tends to give a more critical judgment, saying "Could be better" instead of a larger disagreement like "Yes" vs "No."\\}
\begin{tabular}{lccc}
\hline
Category     & Agree          & Disagree  & Yes vs Could be better \\
\hline
Relevance    & 97/154 (62.99\%) & 57/154 (37.01\%) & 51/57 (89.47\%) \\
Entailment   & 15/20 (75.00\%)  & 5/20 (25.00\%) & 2/5 (40.00\%) \\
Accessibility& 56/84 (66.67\%)  & 28/84 (33.33\%) & 22/28 (78.57\%) \\
\hline
\end{tabular}
\label{tab:gpt_human_agreement}
\end{table}

The moderate 62.99\% agreement in WildfireGPT relevance evaluations is largely due to nuanced differences rather than fundamental disagreements, often due to the data availability issues since retrieved data may not exactly match user-specified timelines or locations, whereas experts were more lenient, acknowledging these constraints. 

In most entailment disagreements, domain experts found responses satisfactory ("Yes"), while GPT-4 flagged them as needing improvement ("Could be better" or "No"). These discrepancies often stemmed from WildfireGPT extrapolating beyond available data. For example, based on FWI data showing minimal change over time, WildfireGPT described wildfire risk as "relatively stable";  GPT-4 criticized this, arguing that it can be misleading and that there are possibilities for wildfire risk to increase. In another case, when asked about fire controls for cherry trees, the retrieved literature did not directly address this specific species but rather the same family of trees in general; experts found this reasonable given the limitations of the available literature, but GPT-4 criticized it as unsupported. Experts appeared more forgiving of such extrapolations, likely because of the experts filling in gaps with their own background knowledge.

Accessibility disagreements highlighted differing standards: GPT-4 recommended simpler wording or more detailed explanations. However, experts' satisfaction varied with their specific expertise and familiarity with the topic. For example, In areas of deep knowledge they may prefer concise, technical responses, while in less familiar areas they require more detailed explanations. GPT-4’s critiques do not fully account for this spectrum of expertise.

In summary, the results reveal moderate agreement between GPT-4 and human evaluations. While GPT-4 can provide scalable and cost-effective evaluations, it may not fully capture the nuanced expectations and information needs of domain experts, leading to discrepancies in relevance, entailment, and accessibility assessments.

%% file: 4.discussion.tex
\edit{This section highlights the core contributions and insights derived from the design and evaluation of WildfireGPT. It focuses on the system’s strengths, particularly its human-centered personalization capabilities, and evaluates how integrating user context enhances the relevance and utility of LLM-generated recommendations. Additionally, this section reflects on the broader implications for domain-specific LLM adaptation and outlines promising directions for future research in AI-assisted climate hazard decision-making.}

\subsection*{Human-Centered Personalization: A Key Strength of WildfireGPT}

Prior efforts to leverage LLMs for climate action have primarily focused on generating factual responses to generic, one-off questions \cite{vaghefi2023chatclimate, thulke2024climategpt, pirozelli2024benchmarks, hsu2024evaluating, cao2024llm}. However, this approach can fail to capture the nuanced expertise required in complex domains, since LLMs tend to provide hasty, generic responses without considering the user's specific context and needs \cite{chen2023challenges, hager2024evaluation, li2024mediq, wang2024learning}.

In contrast, WildfireGPT's key strength lies in its ability to deliver personalized, context-aware recommendations. By integrating user-specific information through its \edit{user profile agent} and generating transparent action plans \edit{through its planning agent}, WildfireGPT tailors its responses to the user's background, location, and specific concerns about natural hazards. \edit{This personalization capability was systematically validated through our ablation study demonstrating how the same query ("Develop recommendations to enhance forest resilience against wildfires") yields distinctly different literature retrieval strategies and recommendations when filtered through various professional lenses. The effectiveness of this approach is quantitatively confirmed by the high relevance and accessibility scores reported,} where domain experts consistently rated the system's recommendations as highly pertinent, not only to their questions, but also to the contexts of their concerns in the \edit{natural hazard} domain. The alignment between user needs and system outputs is a critical advantage of WildfireGPT, ensuring that the information provided is not only scientifically accurate but also practically useful for professionals working in wildfire management. 

Future research should build on the approach demonstrated by WildfireGPT, exploring methods for context-aware domain adaptation of LLMs. By developing techniques to efficiently integrate domain-specific knowledge and problem context into LLM training and inference, researchers can create AI systems that can bridge the critical gap between complex \edit{natural hazard} information and actionable, site-specific strategies for a wide range of related challenges \cite{chaiir2023using, adshead2023mission}. 

\subsection*{Transforming Localized Data into Actionable Insights}

Natural-hazard-informed decision-making requires the integration of multiple location-specific data sources and scientific findings to capture its multifaceted nature \cite{kirchhoff2013actionable, tomaszewski2020geographic, vasquez2021decision, abid2023present}. However, navigating and synthesizing this information can be cumbersome for professionals; using Geographic Information Systems software is time-consuming \citep{kurniawan2023understanding}, writing custom data analysis code requires specialized skills not always available in these roles, and science literature is complex to comprehend \citep{adshead2023mission}. Prior attempts to integrate LLMs for this task have either limited themselves to scientific reports \cite{vaghefi2023chatclimate, toetzke2023leveraging, stagnol2023answering, hsu2024evaluating, thulke2024climategpt}, which can lack local context, or focused solely on data analysis  \cite{kraus2023enhancing,cao2024llm}, missing the transfer of scientific insights for recommendation development. 

\edit{Unlike baseline alternatives such as ChatClimate and Perplexity AI, which demonstrated significant limitations in geographical precision and data accuracy, respectively,} WildfireGPT addresses these challenges by integrating location-specific projections and observational data, socioeconomic indicators, and scientific literature  from trusted sources to deliver an \edit{extensive} analysis of wildfire risks. \edit{WildfireGPT consistently outperforms alternatives in data provision, location specificity, and data accuracy, while maintaining high contextual relevance through literature that shares ecological or geographical characteristics with specified locations. When confronted with data limitations, WildfireGPT demonstrates methodological adaptation by explicitly acknowledging constraints and proposing alternative analytical resources rather than defaulting to geographical generalizations.} A prime example of WildfireGPT's capabilities is shown in Figure \ref{box:Ecosystem_Fire_Management}, where WildfireGPT synthesizes insights from analyzing fire weather indices, historical wildfire data, and scientific literature to provide suggestions on the timing for controlled burns to mitigate wildfire risks.

As natural hazards intensify, the need for tools that can rapidly synthesize complex data into actionable strategies becomes increasingly critical \cite{kirchhoff2013actionable, adshead2023mission, ali2023climate}. The framework and design principles behind WildfireGPT are general and readily adaptable to more data sources. For example, future development could explore the integration of real-time sensor data and community-reported information, creating a more dynamic, responsive system for wildfire risk assessment \cite{ahmed2020forest, abid2023present}. Moreover, more extensive data integration could enable WildfireGPT to provide insights into other climate hazards, such as heatwaves \citep{lopez2018early} and floods \cite{hirabayashi2013global}. Furthermore, as the integration of data sources grows more complex, going beyond text-only models to incorporate multimodality and code generation for more nuanced analysis and flexibility could be a promising direction for future research \cite{ma2024weatherqa, kadiyala2024comprehensive, jiang2024multi}.

\subsection*{Systematic Evaluation of Domain-Specific Conversational Tools}

\edit{Evaluating conversational systems in the natural hazard domain presents unique challenges beyond conventional lexical similarity metrics such as BLEU and ROUGE, which correlate poorly with expert assessment of domain-specific content generation.} Moreover, while existing benchmarking frameworks with well-defined, one-off questions can provide a starting point for assessing AI-generated responses \cite{pirozelli2024benchmarks, hsu2024evaluating,bulian2023assessing}, they fail to capture the nuanced, multifaceted nature of real-world decision-making. \edit{Our case studies highlight that effective climate decision-making is inherently interactive, requiring systems to engage users in ongoing dialogue while synthesizing information from diverse data sources and literature in a personalized manner tailored to their specific professional context and needs.}

\edit{To address these challenges, we developed a three-stage evaluation framework that prioritizes expert judgment over purely automated metrics. First, we conducted a comparative evaluation against ChatClimate and Perplexity AI, assessing data retrieval and evidence-based response capabilities. Second, we performed an ablation study examining how different levels of user profile specificity impact response relevance and actionability across five distinct professional profiles. Third, we evaluated actual responses through domain expert assessment using standardized criteria for relevance, entailment, and accessibility.} This comprehensive approach captures the practical applicability and ease of understanding of the information provided. Future research can build on this framework and explore additional evaluation metrics that capture the real-world utility of LLMs in \edit{natural hazard} decision-making scenarios. In our case studies, domain experts noted instances where WildfireGPT's recommendations aligned with their existing knowledge. However, quantifying this alignment and its impact on decision-making remains an open research question. Going beyond this, future metrics can assess the system's ability to accelerate decision-making processes, improve decision quality, or introduce users to new resources and insights.

Another critical direction is to develop more sophisticated automated evaluation techniques to assess the utility of AI tools in \edit{natural hazard} decision-making at scale. Our preliminary efforts to incorporate LLM-as-a-judge highlight the potential of using AI systems to evaluate AI-generated responses. However, the moderate level of agreement between LLM-as-a-judge and human evaluators suggests that LLM-as-a-judge currently struggles to capture the nuanced understanding of data limitations and context-specific, utility-driven considerations that human evaluators bring to the table. \edit{A promising approach is prediction-powered inference \cite{angelopoulos2023prediction}, which enables valid statistical inference by supplementing small-scale expert-labeled datasets with machine learning predictions without making assumptions about the underlying algorithms. This framework could quantify uncertainty in model-based evaluations and provide confidence intervals that narrow with more accurate predictions, potentially enabling more efficient and rigorous assessment of LLM-based judges against limited human evaluation data. This technique has particular relevance for LLM evaluation, as demonstrated in recent work that constructs statistically valid rank sets from a combination of human and LLM-provided pairwise comparisons \cite{chatzi2024prediction}. This approach could be particularly valuable for evaluating WildfireGPT, allowing us to make reliable conclusions about system performance by combining limited expert feedback with more abundant GPT-based assessments.} Future research may explore approaches such as fine-tuning on domain-specific data \cite{dettmers2024qlora}, leveraging reinforcement learning techniques \cite{carta2023grounding}, and employing more sophisticated prompt engineering \cite{lester2021power, guo2023connecting}. 

\subsection*{Opportunities for Improvement}

\paragraph*{Enhancing Semantic and Spatial Information Retrieval through Integrated Knowledge Frameworks} \edit{To deliver precise, location-aware results, we observe three areas for improvement in semantic processing and geographic contextualization. First, as explained in the Methodology section, we apply semantic search using BERT models to match the semantic similarity between the search query for literature and the abstract of each paper in the database \cite{reimers2019sentence} However, in semantic processing, the reliance on general-purpose language models such as BERT leads to mismatches between user queries and retrieved content—for example, suggesting fuel treatment studies for power grid protection inquiries—due to insufficient domain-specific nuance. Implementing a hybrid hierarchical retrieval framework would help address these limitations by interleaving results from sparse retrievers (capable of processing lengthy documents without structural metadata) with dense retrievers (optimized for semantic similarity) \cite{arivazhagan2023hybrid}. Ontology-enhanced semantic search frameworks \cite{naqvi2025enhancing} can also help formalize relationships between concepts and spatial entities, enabling precise differentiation between management strategies with varying degrees of regional appropriateness. Second, geographic contextualization failures arise when the system retrieves studies from ecologically dissimilar regions (e.g., applying West Coast strategies to Virginia) or overlooks location-specific data gaps (e.g., missing RCP 8.5 projections for Denver). Constructing and integrating spatially informed knowledge graphs, such as KnowWhereGraph \cite{zhu2025knowwheregraph}, with its 29 billion RDF triples spanning environmental, hazard, and demographic domains, would help enable explicit spatial reasoning and improve the final response. Third,  the limited scope of the underlying scientific literature database restricts the system’s ability to retrieve relevant studies even when semantic and geographic logic are sound. Moreover, it is worth noting that abstracts might not capture the full details of papers, potentially causing relevant content to be overlooked. While improving search methods helps maximize existing resources, systematically expanding the knowledge base remains essential for comprehensive coverage. This expansion should particularly target underrepresented regions and niche professional needs where current gaps are most acute.}

\paragraph*{Developing Adaptive Communication Frameworks for Diverse Stakeholders}
\edit{The significant variation in information requirements across stakeholder groups—from the technical precision demanded by climate scientists to the contextual clarity needed by urban planners—underscores the necessity for adaptive communication frameworks in wildfire decision-support systems. WildfireGPT already collects critical user context at the outset of interactions, including profession, primary concerns, and scope of inquiry. Leveraging this metadata to dynamically tailor responses—such as adjustment of technical depth, prioritization of recommendations, or presentation format—remains an area for further development. The Human-Robot Teaming Framework with Multi-Modal Language Feedback (HRT-ML) proposed by Liu et al.~\cite{liu2024effect} offers a promising architectural template for this enhancement. Specifically, its dual-module structure—a coordinator for high-level user intent inference (e.g., inferring that a civil engineer needs slope stability insights rather than general recommendations) and a manager for task-specific adaptation (e.g., prioritizing recommendations in improving slope stability)—could be adapted to WildfireGPT’s workflow. By explicitly mapping user-provided context (profession, location, timeframe) to HRT-ML’s tiered support levels, WildfireGPT could proactively modulate responses: for example, suppressing tangential literature for time-constrained emergency responders or augmenting explanations with local regulatory precedents for planners. This promising approach could address current gaps in personalization, such as overly generic advice for homeowners or insufficient technical rigor for scientists, while maintaining the system’s core strengths in data fidelity and location specificity.}

\paragraph*{Limitations of GPT-4 in WildfireGPT}

Our case studies revealed two significant limitations of GPT-4 that can negatively impact user experience: hallucinations in the form of typos and distractions.
Typos can lead to misspelled words or functions, resulting in invalid function calls. These can cause unsuccessful module transitions or data retrieval failures.
Distractions can also cause failed module transitions. For example, in the Community Hazard Mitigation Planning case study, WildfireGPT failed to transition from the user profile module to the planning module after all questions were answered because it hallucinated that the conversation was complete. These limitations often require manual intervention from the facilitator to correct the issues and move the conversation forward. We note that these issues are not unique to WildfireGPT \cite{shi2023large, ji2023survey}, and future improvements should focus on enhancing the language model to mitigate hallucinations. 
Another challenge is the tendency of GPT-4 to generate lengthy and generic outputs, potentially overwhelming users and obscuring key information. This may be attributed to current Reinforcement Learning from Human Feedback (RLHF) techniques that encourage longer outputs \citep{singhal2023long} and the lack of domain expert involvement in the human labeling component of RLHF. The absence of specialized professionals in the feedback loop can lead to responses that fail to capture the nuanced expertise required in complex domains like wildfire management.

%% file: 7.method.tex
An overview of the WildfireGPT system is shown in Figure \ref{fig:overview}. In this section we provide further details of the multi-agent setup within WildfireGPT and the evaluation framework. \edit{We developed a multi-agent RAG \cite{singh2025agentic} system as a collaborative framework designed to streamline complex decision-making tasks by leveraging specialized agents for different stages of the workflow. Each agent in the system has a distinct role. For instance, the user profile agent collects user inputs and refines the task scope, the planning agent formulates actionable plans based on user-defined objectives, and the analyst agent conducts in-depth analyses to provide insights. These agents work collaboratively under the coordination of the task orchestrator agent, which dynamically routes interactions to the appropriate agent in sequence. In this framework the orchestrator does not act as a decision-maker; rather, it facilitates the flow of information by passing outputs from one agent to the next. Additionally, it supports resuming conversations by replicating the thread history into a new thread, facilitating continuity in long-running or multisession interactions. This modular and scalable design allows the system to efficiently handle diverse and evolving tasks while ensuring seamless transitions between agents.}

\paragraph*{Task Orchestrator}
\edit{We designed the task orchestrator agent to serve as the central coordinator in the multi-agent system, ensuring seamless interaction between the user and the specialized agents for efficient task execution. The orchestrator includes key functions to enable its role through the OpenAI function calling API. The \texttt{get\_response} function processes user inputs, invokes the appropriate agent based on the context, and manages outputs or follow-up actions to keep the interaction smooth and efficient. The \texttt{update\_assistant} function allows the orchestrator to dynamically transition between agents, ensuring that the most relevant agent is assigned to handle the current task based on user input and workflow requirements. The \texttt{resume\_conversation} function is designed to restore context in cases of interrupted or paused conversations by replicating the thread's messages into a new thread, allowing for seamless continuation. Together, these functions are designed to empower the task orchestrator agent to manage workflows, prioritize user-centric engagement, and deliver a cohesive experience across various stages of decision-making and analysis.}

\paragraph*{User Profile Agent}

\edit{The user profile agent is designed to collect and refine user-specific inputs for wildfire risk assessment through a structured, interactive, and iterative approach. This agent helps WildfireGPT produce outputs tailored to the unique needs of each user in a later stage of the interaction. To achieve this, we employed prompting techniques to guide users through individualized questions presented one at a time, enabling a systematic completion of a checklist covering personalized information about the user. Once all questions are answered, the system compiles the responses into a structured checklist and presents a summary to the user for verification, ensuring both accuracy and alignment with the user's expectations. The design also incorporates flexibility, allowing users to respond with "I don't know" if they are uncertain about any question, accommodating varying levels of expertise and familiarity with wildfire risk topics. The agent operates in two distinct stages, predefined inquiry and summary verification, working together to provide understanding of the user’s background and concerns.}

\edit{The first stage of the user profile agent in WildfireGPT involves a structured dialogue designed to gather essential information to prepare for generating actionable wildfire risk insights later. This stage begins with five predefined questions that capture critical user details:  profession, concern, location, time, and scope.}

\begin{itemize}
    \item {\bf Profession:} The profession question collects the user’s professional background and expertise. This information will act as a proxy for WildfireGPT to tailor its recommendations and technical depth to the user’s knowledge level, whether  engineer, urban planner, emergency manager, or infrastructure operator. 
    \item {\bf Concern:} The  concern question identifies the user’s key motivations or queries surrounding wildfire risks, such as community safety, infrastructure resilience, or ecological preservation. This information will be used to help the system target the relevant aspects of wildfire management to the user.
    \item {\bf Location:} The location of interest question identifies the geographic area for wildfire risk assessment, allowing users to input locations as place names, geographic coordinates, or descriptive details. When a user specifies a location in natural language, the system is designed to leverage the LLM to convert the input into geographic coordinates (latitude and longitude). To ensure accuracy, we integrated a map-based verification system that enables users to visually confirm or refine their input. The specified coordinate is then  saved to help retrieve structured data and enable robust and geographically grounded wildfire risk assessments.
    \item {\bf Timeframe:} The timeframe question defines the temporal scope of interest, offering options for short-term (1--10 years) mitigation strategies, medium-term (10--30 years) resilience planning, or long-term (30--80+ years) \edit{resilience} adaptation. Historical analysis is also available, including recent (1--10 years), past (10--50 years), or long-term (50+ years) fire patterns. 
    \item {\bf Scope:} The scope of interest question refines the focus of the risk assessment, covering areas such as infrastructure vulnerability, emergency preparedness, ecological impacts, and insurance planning, with guided examples provided for users who are unsure. After gathering responses, WildfireGPT compiles the information into a structured checklist, which is shared with the user for review to ensure accuracy and alignment with their expectations.
\end{itemize}

\edit{Upon completing and confirming the finalized checklist, the agent saves the information and transitions to the planning agent, where WildfireGPT begins generating actionable insights and recommendations. This transition marks the shift from information gathering to solution generation, ensuring that the outputs are tailored to the user's specific needs and the context of their wildfire risk assessment.}

\paragraph*{Planning Agent}

\edit{We designed the planning agent prompt in WildfireGPT to guide the system in creating a structured and user-centered wildfire risk assessment plan. By embedding step-by-step instructions into the prompt, we enable the planning agent to systematically address user concerns, integrate user feedback, and leverage relevant datasets effectively. 
The prompt for the planning agent comprises the following components:}

\begin{itemize}
    \item Step-by-Step Engagement Plan: The prompt instructs the system to create a short, clear plan to engage with the user effectively. This plan includes leveraging the most relevant datasets to address the user’s concerns through data analysis and a literature search. The system is guided to select datasets based on the user's specific needs—such as analyzing trends in wildfire risk, understanding historical patterns, or focusing on immediate mitigation strategies.
    
    \item Feedback Integration: Once the plan is drafted, the system is prompted to pause and share the plan with the user. This ensures that the user is fully informed about the proposed approach and has the opportunity to provide feedback or request modifications. The prompt explicitly instructs the system to ask whether the user would like to include additional information or datasets, while also clarifying that the wildfire analysis is limited to the three available datasets: FWI, long-term fire history records, and recent fire incident data. This step ensures transparency and active collaboration between the system and the user.
    \item Finalization and Execution: After incorporating user feedback and securing the user's agreement, the system transitions to finalizing the plan. The prompt ends with a specific instruction to call the function \texttt{plan\_complete()} with the completed plan, signaling the readiness to move to the execution phase. 
\end{itemize}
\edit{The planning prompt also leverages one-shot learning by including an example plan that act as a demonstration, guiding the model to understand the desired structure and format of the output. In addition to the example, the prompt includes detailed dataset descriptions, outlining the characteristics and scope of the three available wildfire data sources to constrain the model's data selection. The prompt also integrates the user profile for the agent to make informed decisions.}

\edit{During the user interaction, WildfireGPT shares brief descriptions of the available datasets, before presenting the proposed plan. From the user's perspective, this plan comprises three key steps: (1) data retrieval: WildfireGPT identifies the most relevant wildfire datasets from the three available sources and explains the reasoning behind the selection; (2) literature review: WildfireGPT outlines the thematic focus of the literature to be examined; and (3) recommendation development: based on the data analysis and literature review, WildfireGPT develops personalized recommendations.}

\paragraph*{Analyst Agent}
\edit{The analyst agent leverages the RAG framework \cite{lewis2020retrieval, gao2023retrieval} to conduct analyses by integrating diverse data sources and domain knowledge. This agent maintains the summarized user profile and action plan within its prompt context to ensure that WildfireGPT adheres to the plan and addresses user follow-ups, offering to explore specific topics or proceed to the next step. The analyst agent functions as an information processor, ensuring that outputs are grounded in scientific research articles and wildfire data. Its structured approach enables it to process volumes of information efficiently while maintaining accuracy and relevance. The analysis consists of three main stages:}
\begin{itemize}
    \item {\bf Wildfire data retrieval and visualization}: The data retrieval and visualization functionality in our system is designed to assess wildfire risk by leveraging geospatial data and fire trends. The process begins with location-based data retrieval, based on the location information collected by the user profile agent. For all datasets, data within a 36-kilometer radius of the specified coordinate is retrieved and presented. The results are visualized through interactive maps, tables, and statistical plots. The map-based visualization employs GeoJsonLayer rendering to display the data points. A pin-layer visualization is also integrated to mark the exact user-specified location. To facilitate decision-making, the system dynamically generates a structured report summarizing the statistics.

    {\em Recent Wildfire Incident Record Data:} Historical wildfire locations (2015--2023) are visualized as individual red markers on the GeoJsonLayer, with each incident precisely geolocated within a 36-kilometer analysis radius. In addition, two complementary line plots are presented for temporal analysis: an annual frequency plot revealing incident patterns over the years and a monthly distribution highlighting seasonal variations. The report contextualizes total incident frequency per year and aggregated monthly distributions into risk assessments, analyzing temporal clustering patterns to identify periods of elevated fire activity.

    {\em Tree-Ring and Sediment-Based Fire History Record Data:} The system retrieves and ranks the three nearest paleofire study locations using a geodesic distance algorithm, integrating site-specific metadata from the International Multiproxy Paleofire Database. Each identified site is presented as an individual red marker on the GeoJsonLayer. Additionally, for each site, the system provides the user with the research metadata including site names, precise coordinates, and associated publication records.

    {\em Fire Weather Index Data:} Unlike the other data sources, the FWI data is structured on a grid system and referenced by using Crossmodel indices. Thus, we developed a process to map the user-specified geographic coordinates to the corresponding Crossmodel reference. This process begins by transforming the latitude and longitude into a spatial reference system compatible with the grid model. The geographic coordinates, initially expressed in degrees, are converted into the coordinate reference system used by the FWI data, typically employing projected coordinates for spatial accuracy. Using spatial operations such as buffering and intersecting, the transformed geographic point is compared against the grid cells in the model. The system retrieves the associated Crossmodel indices if they fall within a radius of 36 kilometers of the specified coordinate. This mapping establishes a direct connection between the user-defined location and the corresponding grid cell in the FWI data. By designing this mapping mechanism, we ensured that WildfireGPT can seamlessly integrate user-specified locations with grid-based FWI data, providing precise and scientifically accurate insights tailored to the specified area. The retrieval process integrates historical (1995--2004), mid-century (2045--2054), and end-of-century (2085--2094) projections to offer a temporal perspective on fire weather trends. The retrieved FWI values are aggregated and analyzed to compute mean wildfire indices and standard deviations across seasons, ensuring a statistically robust representation of fire risk. The system then categorizes FWI values into six risk classes (low, medium, high, very high, extreme, and very extreme) based on the Canadian Forest Fire Weather Index classification system \cite{vanwagner1987development}. The GeoJsonLayer displays fire weather indices across spatial grids, with each grid cell color-coded according to its FWI classification. Additionally, the module provides a table presenting the seasonal comparisons of historical, mid-century, and end-of-century projections, enabling users to track long-term fire risk evolution. Lastly, the module is prompted to provide \edit{a} structured summary report detailing past and projected wildfire risks in the region.
    
    {\em Census Data for Socioeconomic Analysis:} If location-specific wildfire data is retrieved, users are informed of an additional analysis capability: demographic and socioeconomic impact assessments from census datasets. 
    This can help provide an understanding of wildfire risks beyond environmental factors, by incorporating population vulnerabilities, economic conditions, and housing density. If wildfire data is not available, the system shares a prewritten cautionary message, emphasizing the preliminary nature of its recommendations and encouraging users to seek further investigation or expert advice before implementing significant changes. 
    
    The census data retrieval functionality extracts demographic and socioeconomic statistics for regions surrounding a user-specified location, with each census block group data retrieved from the American Community Survey (ACS5) dataset. Key indicators such as total population, poverty distribution, and housing unit counts are aggregated to generate a demographic summary, highlighting the number of individuals below the poverty line and those with income less than half the poverty threshold. Each census block group is visualized on the GeoJsonLayer for interactive spatial understanding.
    Additionally, a structured table presents aggregated population and housing statistics, allowing users to explore socioeconomic factors in detail. 
    
    By integrating real-time geospatial data analysis, interactive visualization, and RAG techniques, WildfireGPT ensures that wildfire risk assessments are data-driven, context-aware, and easily interpretable for end users.
    \item {\bf Literature retrieval}: For scientific literature, WildfireGPT queries a corpus developed by Argonne National Laboratory for the CIACC tool \citep{mallick2023analyzing, mallick2024analyzing}. All abstracts of the papers in this corpus are pre-embedded using the \texttt{all-MiniLM-L6-v2 embedder} from the SentenceTransformers library \citep{reimers2019sentence} and stored in a Faiss vector store \citep{johnson2019billion}. The literature search query is processed by the same embedding model and undergoes a k-nearest neighbor search, identifying the top k-most relevant abstract based on cosine similarity between the query vector and stored paper embeddings in the FAISS vector database. To enhance the reliability of the retrieved results, the system attempts to validate DOIs (Digital Object Identifiers) by cross-referencing them with CrossRef metadata. This validation process includes retrieving DOIs based on the paper title, checking for title and author consistency, and computing a similarity score to confirm accuracy. If discrepancies such as title mismatches or incorrect author attributions are detected, the DOI is discarded to maintain data integrity and prevent misinformation.  The final output is structured as a ranked list of the three most relevant papers, displaying each paper’s title, authors, publication year, and abstract, with a direct DOI link if verified. This integration of FAISS-based vector search, semantic similarity ranking, and DOI validation ensures the provision of high-quality, research-backed insights for scientists, policymakers, and decision-makers in wildfire risk assessment and mitigation planning.

    \item {\bf Generation of tailored recommendations}: 
    To generate tailored recommendations, the system constructs an augmented prompt that incorporates context from the ongoing conversation to provide actionable recommendations. If additional clarification or exploration is required, the system iterates the retrieval and response process, continuously refining the output to address the user’s evolving needs. In this way, the system contextualizes data retrieval and recommendation generation based on the user’s profession, ensuring that risk assessments, mitigation strategies, and policy suggestions align with their specific safety priorities and operational responsibilities.
\end{itemize}

\subsection*{Details of the Evaluation Framework}

\paragraph*{Personalization}
\edit{We carry out a two-phase ablation study to evaluate the impact of user profile granularity on  WildfireGPT's responses. In the first phase, we use the prompt “Develop recommendations to enhance forest resilience against wildfires” to elicit recommendations from WildfireGPT, with three levels of user profile details: }
\begin{enumerate}
    \item \textbf{Generic}: No contextual details.
    \item \textbf{Location + Timeline}: Added geospatial (Covington, VA; 37.7935°N, 79.9939°W) and temporal (5--10 year implementation window) parameters.
    \item \textbf{Full Profile}: Integrated profession-specific attributes (e.g., \textit{power grid manager}), operational concerns (e.g., grid resilience, transmission line clearance), and scope (e.g., infrastructure protection).
\end{enumerate}

\edit{In the second phase, we hold location and timeline constant and test five distinct professional profiles (\textit{homeowner, civil engineer, ecologist, emergency manager, power grid manager}). For each profile, we modify three variables---profession, concern, and scop---in the following template user profile:}
\fbox{%
  \begin{minipage}{0.9\textwidth}%
    \texttt{%
      \begin{enumerate}
        \item Profession: \{profession\} in Virginia.
        \item Concern: Managing the forest, keeping it healthy, while \{concern\}, and protecting properties from potential wildfires.
        \item Location: Near Covington, VA with Latitude 37.7935 and Longitude -79.9939.
        \item Time: Recommendations to be implemented within the next 5 to 10 years.
        \item Scope: Management of the forest and properties to maximize \{scope\}, and protect against potential wildfires.
      \end{enumerate}
    }%
  \end{minipage}%
}

\paragraph*{LLM-as-a-Judge}
We prompt, in a zero-shot manner, GPT-4 to assess the relevance, entailment, and accessibility of the WildfireGPT responses using the same questions posed to human experts in Table~\ref{tab:human_eval_criteria} with minor modifications such as adjusting the pronouns---replacing "\textit{my}" with "\textit{the}" and "\textit{your}" with "\textit{the user's}." Additionally, GPT-4 as  judge is instructed to provide one of four possible judgments---Yes, No, Could be better, or Not applicable---and to offer explanations for each assessment. For all evaluations, it receives the user profile and concerns, the user's previous queries, retrieved scientific literature and data, and the model's responses as inputs.

%% file: 5.data.tex
The FWI projections from ClimRR are available for download at \url{https://anl.app.box.com/s/hmkkgkrkzxxocfe9kpgrzk2gfc4gizp8}. Wildland Fire Incident Locations data can be accessed at \url{https://data-nifc.opendata.arcgis.com/datasets/nifc::wildland-fire-incident-locations/about}. The North American Tree-Ring Fire Scar Synthesis  dataset can be downloaded from \url{https://www.ncei.noaa.gov/access/metadata/landing-page/bin/iso?id=noaa-fire-34853}. Census ACS5 data is accessible via the Python API package at \url{https://github.com/Census-ACS/census}. The scientific literature data developed by CIACC can be downloaded from \url{https://anl.box.com/s/b4m2mnt5wa4z9l71qioz05cb5qpduj2j}. All data is available for download at \url{https://anl.box.com/s/wm888zovyapyou1txae7g75ghpc7sxre}.

%% file: 6.code.tex
The code for WildfireGPT is publicly available on GitHub at \url{https://github.com/Xieyangxinyu/WildfireGPT}. The repository contains all the necessary components for reproducing the functionality of WildfireGPT, including the prompts used to query GPT-4, functions for identifying the location of interest (through a pinpoint mapping tool), visualizing data, and enabling transitions between modules based on the user's input. These functions are integrated with function calling support provided by the OpenAI Assistant API \citep{openai2023assistant}. The software is implemented by using Python 3.11, and the front-end user interface is built as a Streamlit-based web application \citep{streamlit}.

%% file: 8.appendix.tex
\section*{Supplementary Materials}

\subsection*{Overview of the WildfireGPT User Experience}

\begin{figure}[hbt!]
    \centering
    \includegraphics[width=0.32\linewidth]{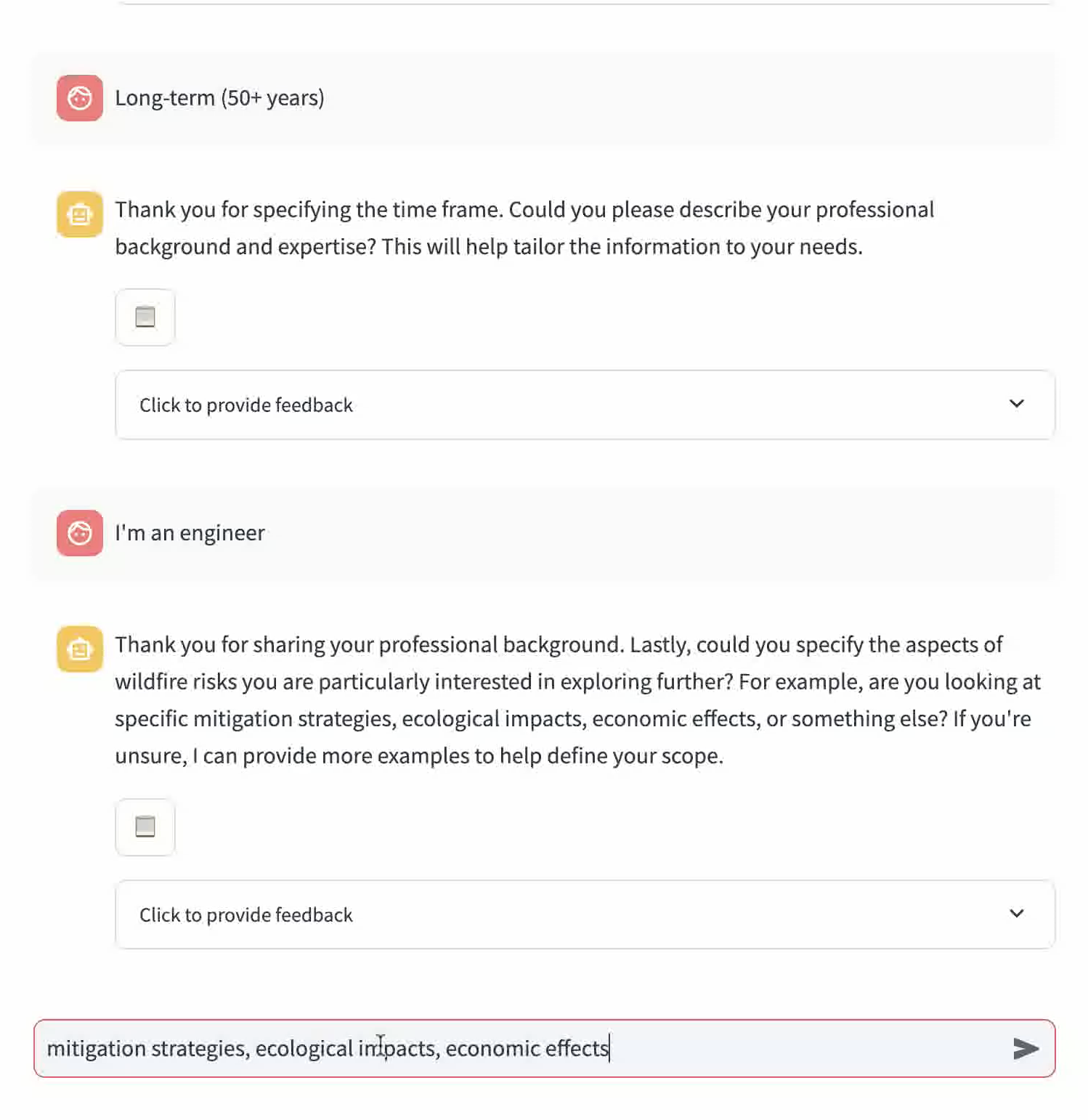}
    \includegraphics[width=0.32\linewidth]{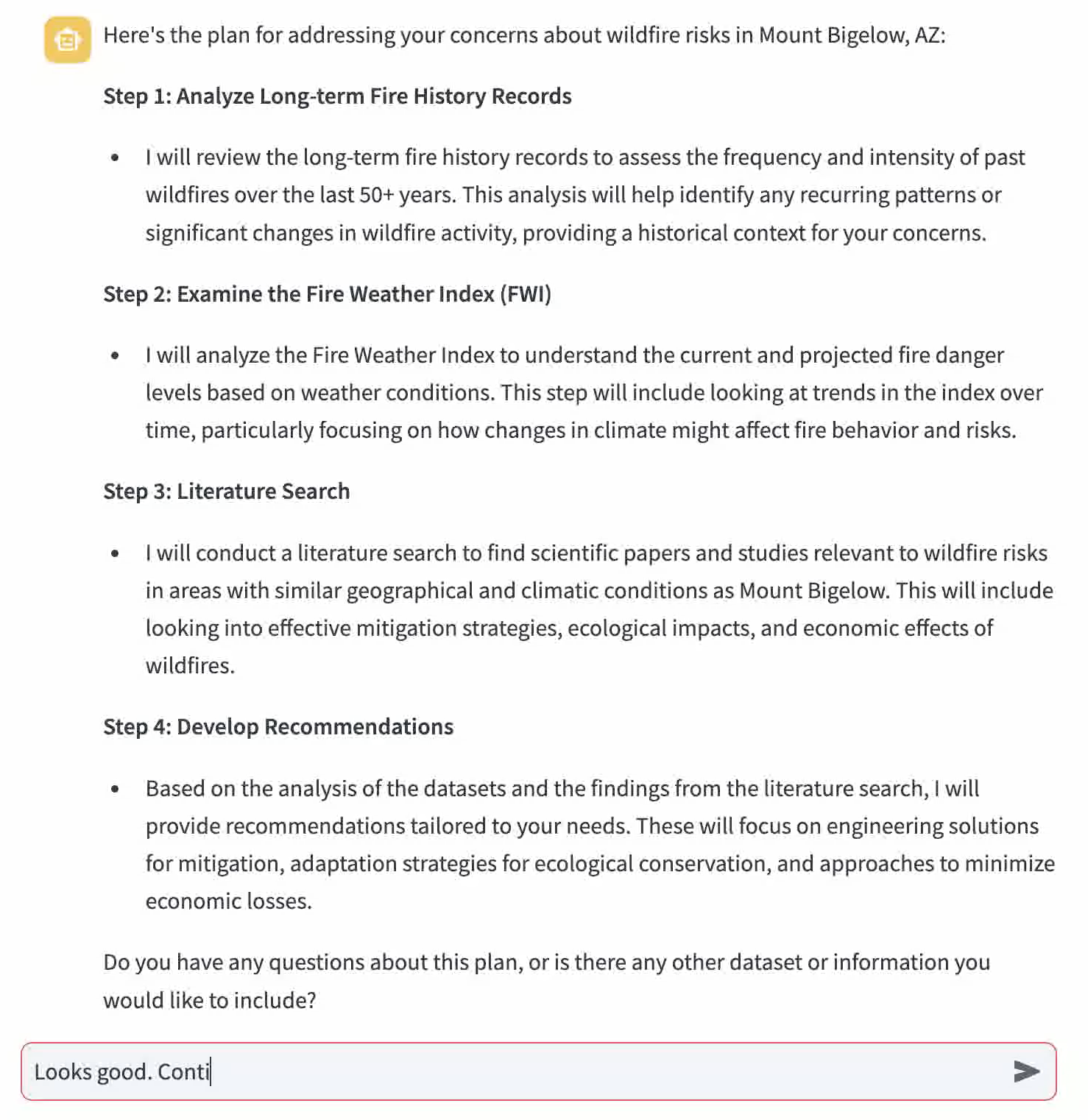}
    \includegraphics[width=0.32\linewidth]{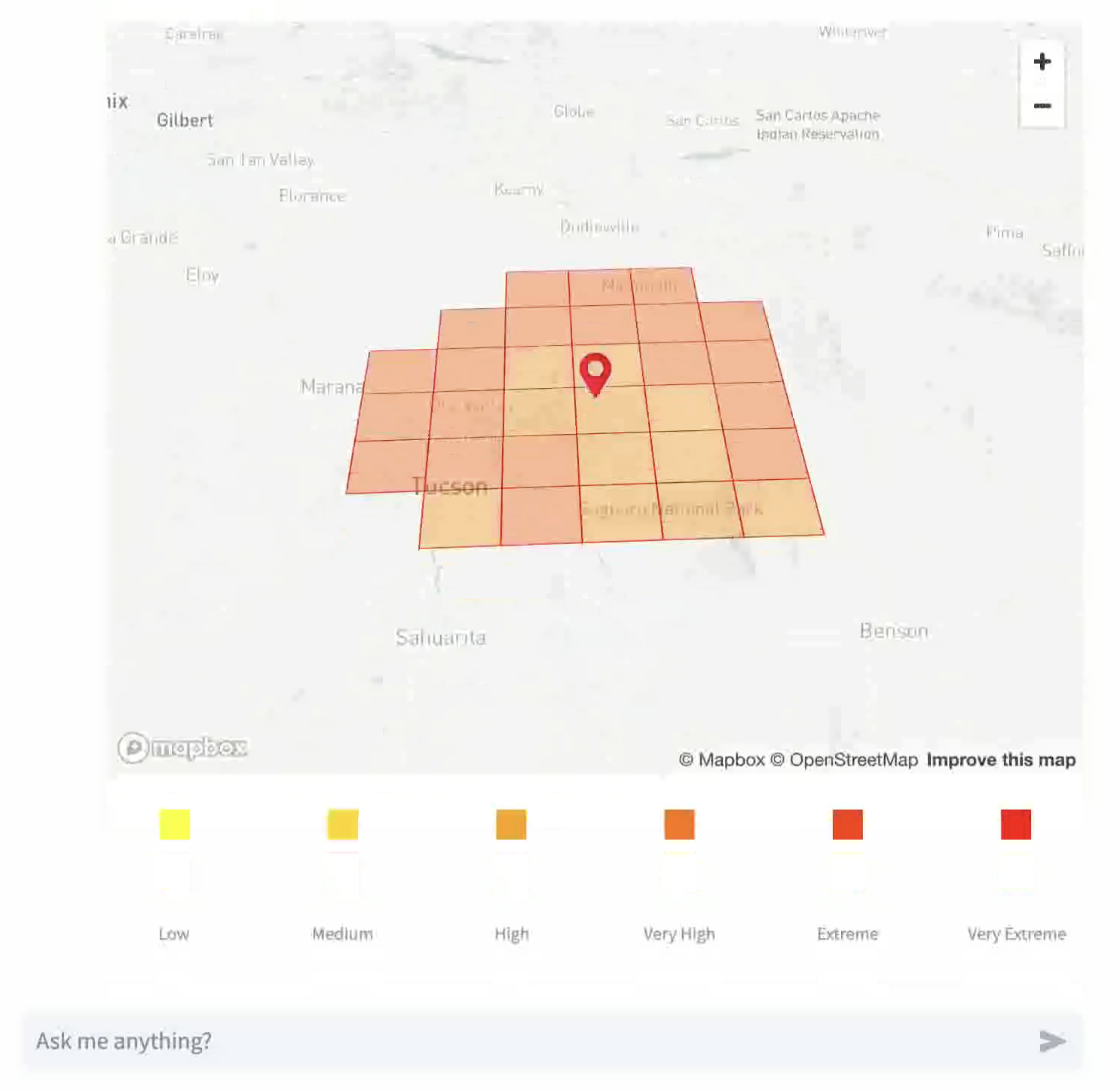}
    \includegraphics[width=0.32\linewidth]{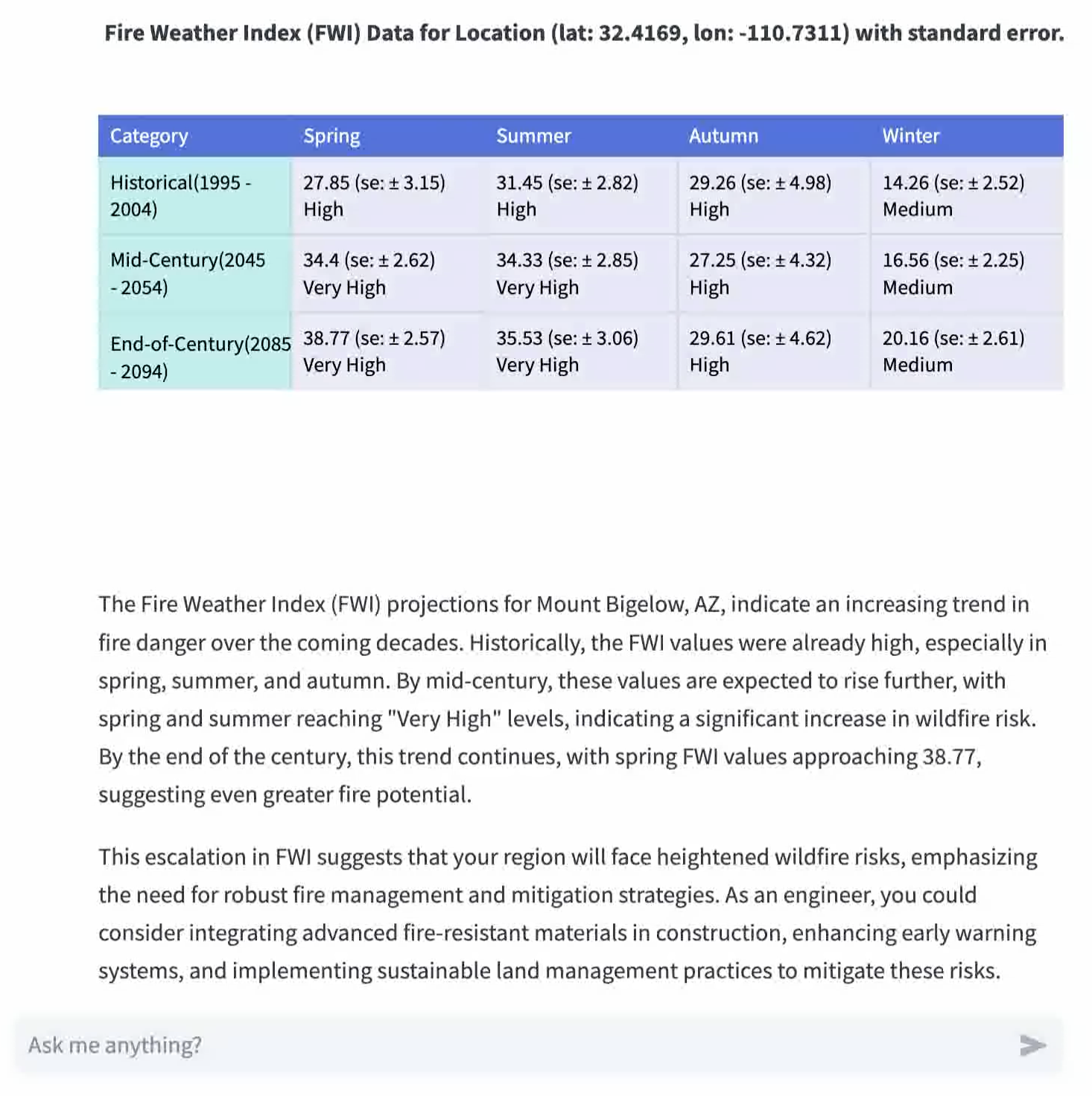}
    \includegraphics[width=0.32\linewidth]{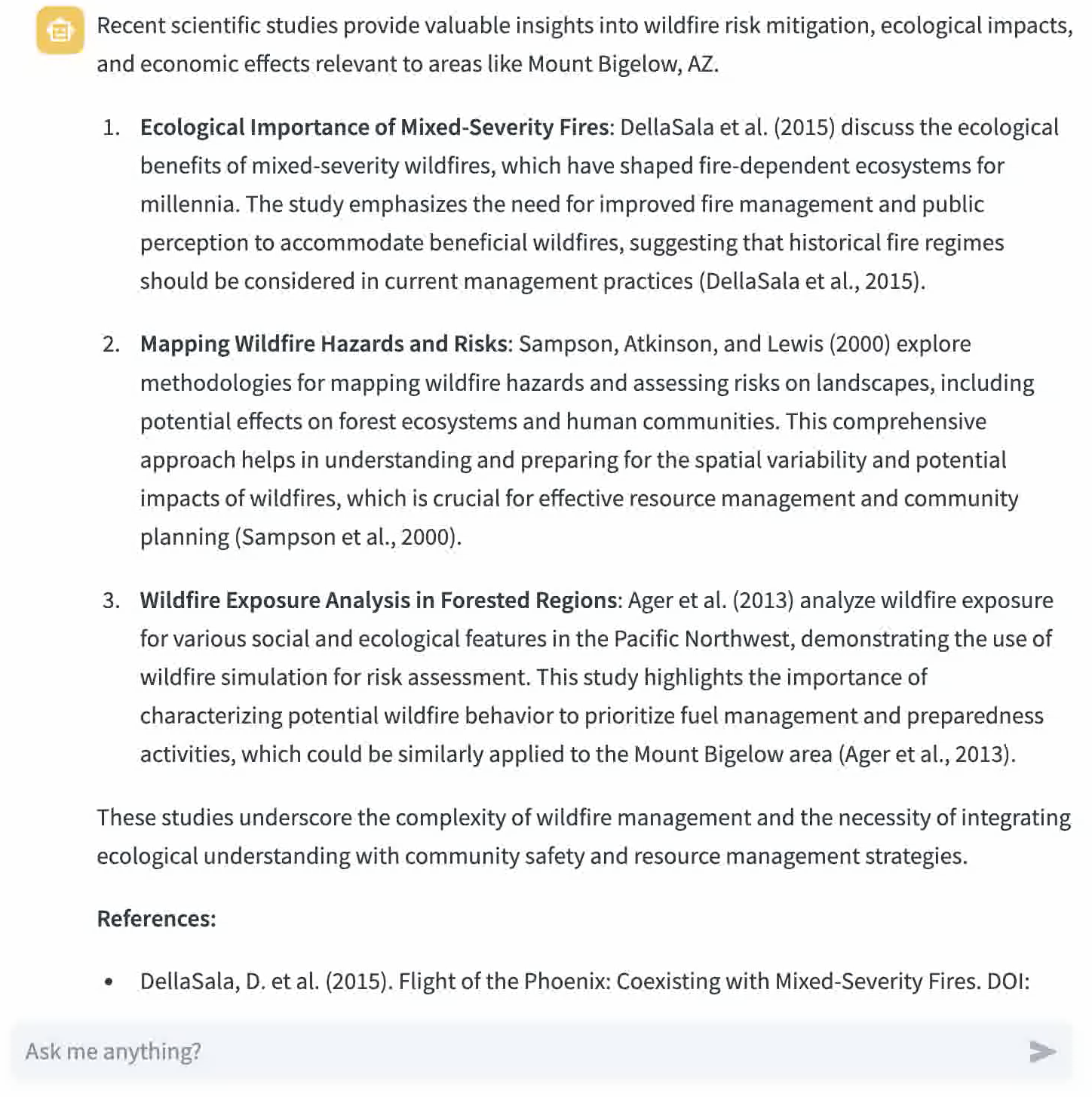}
    \includegraphics[width=0.32\linewidth]{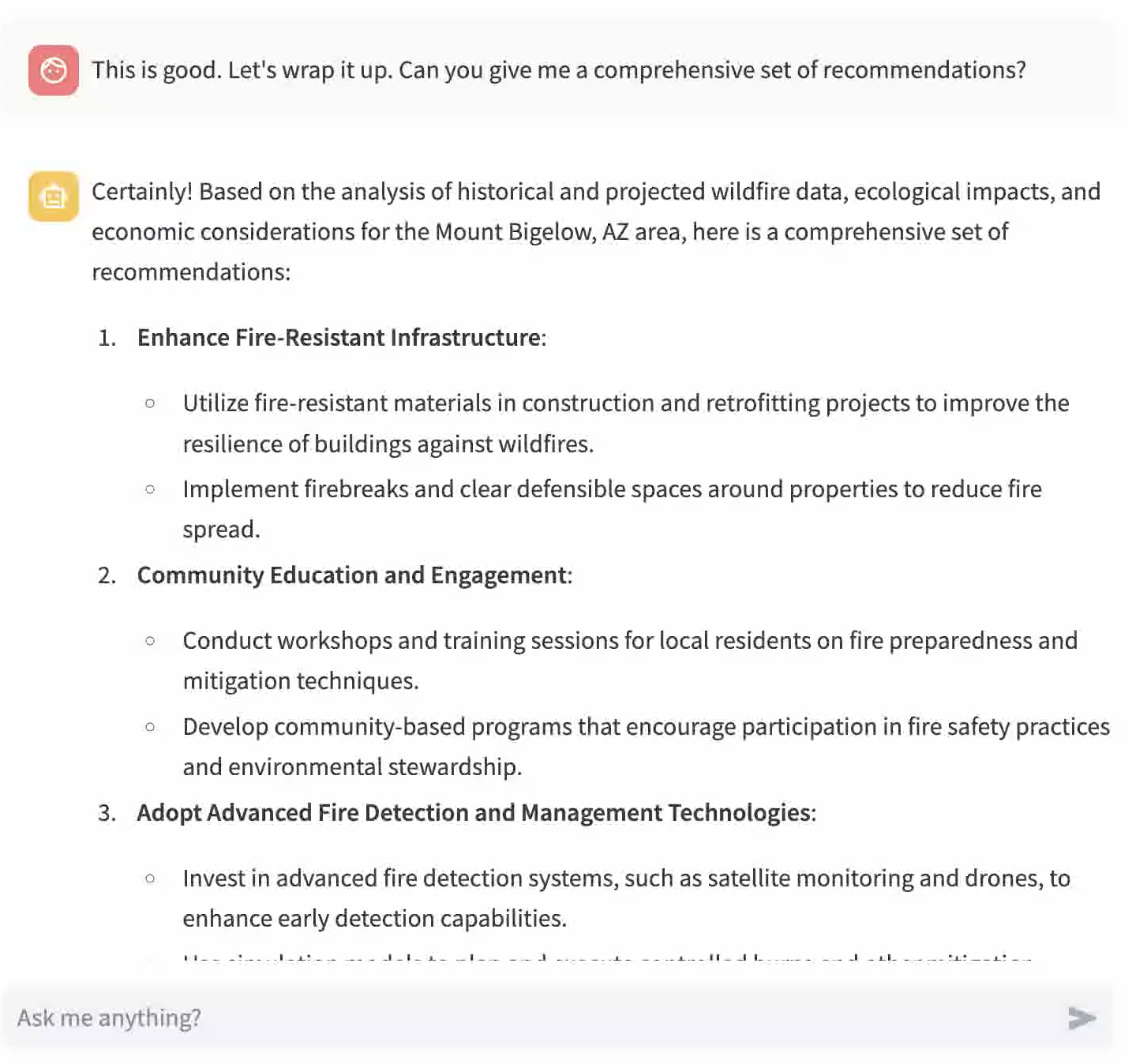}
    \caption{Overview of the WildfireGPT user experience. The screenshots are taken from one of the case studies themed Comprehensive Wildfire Impact. The user profile agent (top left) engages the user in a conversation to understand the user's background and concerns. The planning agent (top middle) generates a tailored analysis plan based on the user's profile. The analyst agent then executes the plan, analyzing Fire Weather Index data (top right and bottom left), conducting a literature review (bottom middle), and generating personalized recommendations (bottom right) to address the user's wildfire risk concerns.}
\label{fig:wildfiregpt-overview}
\end{figure}

Supplementary Figure \ref{fig:wildfiregpt-overview} provides an overview of the user experience as WildfireGPT transitions through the three agents, with screenshots from one of the case studies themed Comprehensive Wildfire Impact.

\subsection*{Comparison against Baseline Models}

To evaluate the effectiveness of data retrieval and evidence-based query response capabilities, we conducted systematic comparisons between WildfireGPT and two baseline models: ChatClimate and Perplexity AI. Our experimental framework comprised two distinct analytical modules drawn from our case studies (Tables \ref{table:prompts-data} and \ref{table:prompts-QA}): (i) data retrieval and analysis tasks, which examined each model's capacity to process and interpret location-specific wildfire data across diverse geographical and temporal scales, and (ii) evidence-based query responses, which assessed the models' ability to provide evidence-based answers to domain-specific questions. The evaluation metrics for data analysis encompass data provision (successful retrieval of relevant data), location specificity (geographical precision of retrieved data), and data accuracy (alignment between retrieved data and prompt requirements). Similarly, question-answering response assessment examined external knowledge integration, source authority, citation practices, and contextual relevance. Below is an example of the responses provided in the by each model. The full set of responses can be accessed in the provided GitHub repository. 

\begin{table}[h!]
\centering
\begin{tabular}{|p{0.35\linewidth}|p{0.65\linewidth}|}
\hline
\textbf{Case Study} & \textbf{Prompt} \\
\hline
Climate Impact on U.S. Wildfire Occurrences and Intensity & Analyze future climate data to understand potential trends in fire danger in Boston, Massachusetts, USA \\
\hline
Comprehensive Wildfire Impact & Review the long-term fire history records to assess the frequency and intensity of past wildfires over the last 50+ years in Mount Bigelow, AZ \\
\hline
Current Wildfire Risk Analysis & Analyze demographic and socio-economic profiles of the Chico area in California to understand which groups are most vulnerable in wildfire scenarios \\
\hline
Ecosystem Fire Management & Analyze the Long Term Fire History Records to assess wildfire events over the past 150 years in the Naperville region \\
\hline
Hazard Mitigation Planning & Analyze the recent fire incident data for Mora County, reviewing wildfire frequency, intensity, and timing from 2015 to 2023 \\
\hline
Infrastructure Wildfire Risk & Examine climate models and projections under the RCP 8.5 scenario to forecast potential future wildfire risks in the medium-term (10--30 years) in Denver, Colorado \\
\hline
Post-Wildfire Public Safety and Environmental Restoration & Analyze the recent fire incident data from the Wildland Fire Interagency Geospatial Services Group to assess the frequency, intensity, and locations of recent wildfires in Sangre de Cristo Mountains \\
\hline
Private Property Protection & Analyze the recent fire incident data from the last five years in Covington, VA \\
\hline
Urban Wildfire Mitigation & Analyze the FWI focusing on both current trends and projections for the mid-century (2045--2054) to assess the potential increase in fire danger due to climate change in Beaverton, Oregon \\
\hline
Wildland Urban Interface Impact on Housing and Water Resources & Analyze projected wildfire risk data for the mid-century period (2045--2054) to understand the potential trends in fire danger in Las Vegas, NM \\
\hline
\end{tabular}
\caption{Prompts Used in for Modular Comparison in Data Retrieval}
\label{table:prompts-data}
\end{table}

\begin{table}[h!]
\centering
\begin{tabular}{|p{0.35\linewidth}|p{0.65\linewidth}|}
\hline
\textbf{Case Study} & \textbf{Prompt} \\
\hline
Climate Impact on U.S. Wildfire Occurrences and Intensity & Is FWI a robust estimate of the fire danger risk? \\
\hline
Comprehensive Wildfire Impact & What are the wildfire mitigation strategies, ecological impacts, and economic eﬀects in forested areas similar to Mount Bigelow, AZ? \\
\hline
Current Wildfire Risk Analysis & Please provide sources on how to obtain data for Vegetation Type and Density, Previous Burn History, Human Activity Levels, Topography, Weather Conditions, Land Management Practices, Infrastructure and Accessibility, Socio-Economic Factors, and Climate Change Indicators \\
\hline
Ecosystem Fire Management & It looks like March is the time folks implement a controlled burn in Naperville, IL. What do you think? \\
\hline
Hazard Mitigation Planning & What universities would make good partners to undertake studies on fire behavior and eﬀective mitigation strategies in Mora County, NM?  \\
\hline
Infrastructure Wildfire Risk & Let's dive deeper into Firebreaks and Buﬀer Zones. I'm interested in learning about approaches to establish firebreaks around the critical infrastructure and vegetation management practices. Can you provide more detail with evidences? \\
\hline
Post-Wildfire Public Safety and Environmental Restoration & Search for scientific literature on eﬀective post-wildfire recovery strategies, focusing on public safety and environmental restoration in Sangre De Cristo Mountains. \\
\hline
Private Property Protection & Is fire common in this region prior to 1900s in Covington, VA? What is the natural fire regime? \\
\hline
Urban Wildfire Mitigation & Are there supporting literature for Buﬀer Zones that communities can implement to manage wildfire risk? \\
\hline
Wildland Urban Interface Impact on Housing and Water Resources & Are there good examples of similar cities to Las Vegas, NM that might experience similar shifts in wildfire risk? What mitigation strategies are those cities implementing? \\
\hline
\end{tabular}
\caption{Prompts Used in for Modular Comparison in Question and Answer}
\label{table:prompts-QA}
\end{table}

\subsubsection*{Examples}
\label{sec: example modular comparison}

\textbf{Case Study:} Wildland Urban Interface Impact on Housing and Water Resources

\textbf{Prompt:} Analyze projected wildfire risk data for the Mid-Century period (2045 - 2054) to understand the potential trends in fire danger in Las Vegas, NM.

\textbf{ChatClimate:} Based on the information provided, the projected wildfire risk data for the Mid-Century period (2045 - 2054) suggests that there will be an increase in fire danger in Las Vegas, NM. Climate change is expected to drive future increases in North American fire weather, particularly in the southwest, with a significant increase in the Fire Weather Index (FWI) before 2050 under the RCP8.5 scenario. This increase includes the frequency of 95th-percentile FWI days, peak seasonal FWI average, fire weather season length, and maximum fire weather index. The fire season across North America is projected to expand dramatically beyond 2°C global warming levels. Additionally, there will be more conducive conditions for lightning-ignited fires in the southeastern USA by mid-century, and warming conditions in Alaska will push temperatures above a threshold for fire danger across Alaska's tundra and boreal forest. Therefore, the trends indicate a heightened risk of wildfires in Las Vegas, NM during the Mid-Century period due to the projected increases in fire danger driven by climate change. (Page: 66, Reference: IPCC\_AR6\_WGI\_Chapter12)

\textbf{Perplexity:} Las Vegas, NM is likely to face increased wildfire risk by mid-century (2045-2054) due to several factors:

\begin{enumerate}
\item Increased potential for very large fires: The risk of very large fires (top 10\% of fires) could increase up to sixfold in parts of the West, including New Mexico, by mid-century (2041-2070) compared to the recent past (1971-2000)[3].

\item Climate change impacts: Warming temperatures, more frequent heat waves, and diminished soil moisture during the dry season are expected to contribute to increased fire danger[3]. These conditions, combined with historic fire suppression practices, may lead to more frequent large fire events.

\item Extended fire seasons: Climate change is projected to both intensify fire-friendly weather conditions and lengthen the season during which very large fires tend to spread[3].

\item Drought and fuel conditions: The combination of warming temperatures and below-normal precipitation in the Southwest may lead to drier fuels and increased fire potential[1].

\item Watershed vulnerability: Las Vegas, NM relies on a single surface water source (the Gallinas River watershed), making it particularly vulnerable to wildfire impacts on its water supply[2]. The 2022 Hermit's Peak/Calf Canyon Fire demonstrated the severe consequences of large-scale fires on the city's water resources.
\end{enumerate}

While specific data for Las Vegas, NM is not provided, the overall trend for the western United States suggests a significant increase in fire danger. To mitigate these risks, it will be crucial for Las Vegas to implement preventive measures, such as watershed management, infrastructure upgrades, and improved emergency response planning.

\paragraph{Citations}:
\begin{enumerate}
\item \url{https://www.nifc.gov/nicc-files/predictive/outlooks/monthly_seasonal_outlook.pdf}
\item \url{https://www.circleofblue.org/2023/world/new-mexicos-largest-fire-wrecked-this-citys-water-source/}
\item \url{https://www.climate.gov/news-features/featured-images/risk-very-large-fires-could-increase-sixfold-mid-century-us}
\item \url{https://www.nifc.gov/nicc-files/predictive/outlooks/NA_Outlook.pdf}
\item \url{https://srs.fs.usda.gov/pubs/gtr/gtr_srs243.pdf}
\item \url{https://pmc.ncbi.nlm.nih.gov/articles/PMC9076366/}
\end{enumerate}

\textbf{Perplexity (when provided with the data file for FWI):} Chose crossmodels "R382C098", "R438C055", "R395C128", "R411C070", "R418C086" in the dataset. However, the correct crossmodels are around "R110C303."

\textbf{WildfireGPT:} The Fire Weather Index (FWI) data for Las Vegas, NM, reveals a concerning trend in fire danger levels over the next few decades. Historically, from 1995 to 2004, the FWI values indicated a medium risk during spring, summer, and autumn, with lower risks in the winter. Specifically:
\begin{itemize}
\item \textbf{Spring:} 13.1 (Medium)
\item \textbf{Summer:} 17.04 (Medium)
\item \textbf{Autumn:} 19.31 (Medium)
\item \textbf{Winter:} 6.98 (Low)
\end{itemize}

Looking ahead to the mid-century period (2045--2054), the projections show a slight increase in the FWI values, especially in the spring:
\begin{itemize}
\item \textbf{Spring:} 17.31 (Medium)
\item \textbf{Summer:} 18.26 (Medium)
\item \textbf{Autumn:} 16.25 (Medium)
\item \textbf{Winter:} 8.49 (Low)
\end{itemize}

By the end of the century (2085--2094), the spring season is expected to experience a significant increase in fire danger:
\begin{itemize}
\item \textbf{Spring:} 23.82 (High)
\item \textbf{Summer:} 20.43 (Medium)
\item \textbf{Autumn:} 20.5 (Medium)
\item \textbf{Winter:} 11.52 (Medium)
\end{itemize}

This trend suggests an increasing risk of wildfires, particularly in the spring, which will necessitate enhanced fire management and urban planning strategies to mitigate risks to housing and water resources. The higher FWI values indicate drier and potentially windier conditions, which can facilitate the rapid spread of wildfires.

Given these projections, it is crucial to integrate robust wildfire risk management strategies into urban development plans for Las Vegas, NM. These should include considerations for fire-resistant building materials, strategic water resource management, and emergency response planning.

[Data Source: Climate Risk \& Resilience Portal (ClimRR)](\url{https://disgeoportal.egs.anl.gov/ClimRR/})

\subsection*{More on Personalization}

Table \ref{tab:phase one full} provides the full details on the comparative analysis across three profile specificity tiers. The full experiment results can be accessed in the provided GitHub repository. 

\begin{table}[htbp]
\centering
\caption{Comparative analysis across three profile specificity tiers. WildfireGPT responses demonstrate progressive domain adaptation from generic principles to stakeholder-specific protocols when provided with increasing user profile granularity. Representative outputs shown for vegetation management, technology implementation, and role-specific interventions demonstrate systematic enhancement of recommendation precision with profile enrichment. Italicized text highlights power-grid-specific recommendations emergent only with comprehensive user profiling.}
\small
\begin{tabular}{|p{0.12\textwidth}|p{0.27\textwidth}|p{0.27\textwidth}|p{0.27\textwidth}|}
\hline
\textbf{Aspect} & \textbf{No Profile Information} & \textbf{+ Location and Timeline} & \textbf{+ Power-Grid Manager Profile} \\
\hline
\textbf{Vegetation Management} & 
``Thinning and Pruning: Reduce forest density through thinning and pruning [...] Controlled Burns: Implement controlled burns to reduce available fuels.'' & 
``Develop and promote fire-adapted ecosystems by incorporating native species [...] Practice selective cutting to reduce fire risk.'' & 
``Establish routine schedules for pruning and removing trees that {\it pose a risk to power lines}. Utilize drones or satellite imagery to monitor vegetation growth for timely intervention.'' \\
\hline
\textbf{Infrastructure} & 
``Enforce fire-resistant materials in construction [...] Maintain and expand firebreaks.'' & 
``Build and maintain natural and artificial firebreaks such as roads [...] or barriers devoid of vegetation.'' & 
``Retrofit existing infrastructure with fire-resistant materials [...] {\it replace wooden poles with those made from steel} or reinforced concrete to enhance fire resilience.'' \\
\hline
\textbf{Technology} & 
``Employ technologies like satellite imaging and drones [...] Use artificial intelligence to predict fire patterns.'' & 
``Invest in enhanced surveillance systems [...] satellite imagery, drones, and remote sensors to detect early signs of fire.'' & 
``Utilize Geographic Information Systems (GIS) and remote sensing [...] Install sensors and advanced monitoring systems along the grid to detect temperature changes, smoke, or increased humidity levels indicating the risk of fire. Use drones for real-time surveillance [...] of {\it power lines, especially in hard-to-reach areas.}'' \\
\hline
\textbf{Role-Specific Actions} & 
``International cooperation [...] Reduce carbon footprints.'' & 
``Encourage Responsible Logging Practices [...] Climate Adaptation Strategies.'' & 
``{\it Expand the Right-of-Way (ROW) clearance around transmission lines} beyond standard regulations [...] Implement an IVM program that combines mechanical, biological, and chemical methods to manage vegetation'' \\
\hline
\textbf{Collaboration} & 
``Foster coordination among forestry services, firefighting agencies, and wildlife organizations [...] International Cooperation: Exchange knowledge [...] with international bodies.'' & 
``Forge partnerships with local universities, environmental organizations, and governmental agencies.'' & 
``Collaborate with local government agencies, environmental organizations, {\it and other utilities} to share resources, information, and strategies for effective forest and wildfire management.'' \\
\hline
\textbf{Policy \& Funding} & 
``Enforce building codes [...] Integrate fire risk into land use planning decisions.'' & 
``Seek federal and state grants aimed at forest conservation [...] fire management.'' & 
``Advocate for local and state policies that support sustainable forest management [...] Seek state and federal grants for implementing advanced technologies [...] in infrastructure resilience and environmental management.'' \\
\hline
\textbf{Community Education} & 
``Train local firefighters and communities in fire suppression techniques and evacuation procedures.'' & 
``Develop community awareness and educational programs [...] Workshops, training sessions, and school curricula can focus on how individuals can reduce wildfire risks through proper land management.'' & 
``Regularly communicate with local communities about maintenance schedules, controlled burns, and emergency procedures to enhance public awareness and cooperation.'' \\

\hline
\end{tabular}
\label{tab:phase one full}
\end{table}

Below we enumerate the plans generated by WildfireGPT for each different user profile in the second phase of our ablation study.

\begin{enumerate}
    \item[Emergency Manager]
    Step 1: I will analyze the Fire Weather Index (FWI) dataset to assess the trends in fire danger over time, focusing particularly on future projections around your area. This will help understand the potential increase in fire risks facing the Covington area in upcoming years, aiding in strategic planning for emergency responses and forest management.\\
    Step 2: I will delve into the Long Term Fire History Records to extract data relevant to the historical wildfire activity around Covington, VA. Analyzing these trends will provide a contextual understanding of the periodicity and severity of wildfires, which is crucial for planning long-term forest management and fire prevention strategies.\\
    Step 3: As your timeframe concerns the next 5 to 10 years, it is critical to also analyze Recent Fire Incident Data. This will highlight the most immediate risks and recent trends in wildfire occurrences. Insights from this will be pivotal in adjusting current practices and preparing immediate response strategies.\\
    Step 4: I will conduct a comprehensive literature search focusing on similar geographical areas with comparable forest compositions and climate to Covington, VA. This research will look into effective forest management practices, successful defendable space creations, evacuation protocols, and property protection measures against wildfires.\\
    Step 5: Based on data analysis and findings from scientific literature, I will provide tailored recommendations. These will address creating or refining defendable spaces, designing evacuation routes, improving property resilience against wildfires, and enhancing overall emergency preparedness and forest health.
    \item[Ecologist]
    Step 1: I will begin by analyzing the Fire Weather Index (FWI) dataset. This will focus on understanding the future trends in fire danger, particularly for the Mid-Century period (2045 - 2054). Analyzing these future projections, and comparing them against the historical period (1995 - 2004), will help us anticipate changes in wildfire risk near your area in Virginia.\\
    Step 2: Next, I will examine the Long term fire history records. This dataset will provide vital insights into the historical patterns and cycles of wildfire activity in the broader North American region over centuries. Understanding these long-term trends will help in assessing the natural fire regimes and their impacts on ecosystems, which is crucial for managing forest landscapes effectively.\\
    Step 3: Concurrently, I will conduct a literature search focused on wildfire management and ecological resilience in forest ecosystems. This will include finding scientific studies that discuss successful strategies for vegetation management, biodiversity preservation, and wildfire prevention in areas with similar ecological and climatic conditions to Covington, VA.\\
    Step 4: Based on the insights obtained from the data analyses and scientific literature, I will develop a set of tailored recommendations. These will be aimed at enhancing the ecological resilience of your forest, improving habitat connectivity, and implementing effective wildfire risk mitigation strategies to protect properties within the next 5 to 10 years.\\
    This plan will help ensure that your forest management strategy aligns well with the evolving dynamics of wildfire risks and ecological demands.
    \item[Civil Engineer]
    Step 1: Analyze the Fire Weather Index (FWI) dataset to measure future fire danger trends, focusing on the Mid-Century period (2045 - 2054). This data will give insight into how fire risk is likely to change and escalate, allowing for forward-thinking preparations.\\
    Step 2: Examine the Long Term Fire History records to understand centuries-long patterns and cycles in wildfire activity. This historical data analysis will contextualize the specific wildfire risks in your region and assist in predicting future trends more accurately.\\
    Step 3: Conduct a thorough literature search of scientific papers focused on wildfire risks and climate change. This search will particularly emphasize studies related to forest management practices, resilience of structures against wildfires, and comparison of geographically similar areas to Covington, VA.\\
    Step 4: Leverage insights from the aforementioned data analyses and literature search to formulate strategic recommendations. These will focus on enhancing forest and property management to improve drainage efficiency, slope stability, and overall resilience against future wildfire threats.\\
    This strategic approach will ensure that all aspects of wildfire risks are comprehensively addressed, aligning with your needs to protect and sustainably manage the forest area around Covington, VA.
    \item[Homeowner]
    Step 1: I will analyze the Fire Weather Index (FWI) dataset to assess the potential trends in fire danger around your location near Covington, VA, for both the Mid-Century (2045 - 2054) and the historical period (1995 - 2004). This will help us understand how the fire danger is expected to evolve over time, which is crucial for long-term forestry and property management planning.\\
    Step 2: I will explore the Long term fire history records to extract and analyze historical wildfire activity patterns in your region. This analysis will focus on identifying any recurring wildfire events and the conditions associated with high fire risks. This information will be vital in predicting future wildfire tendencies and planning effective management practices for your forest.\\
    Step 3: I will conduct a literature search to find scientific studies and successful strategies for managing forests to maintain health, maximize marketable species, and minimize wildfire risks. This will include examining case studies from areas with similar geographical and climatic conditions as near Covington, VA.\\
    Step 4: Based on the compiled data and insights from the literature, I will provide recommendations tailored to your specific concerns and goals. These will include actionable strategies for forest management, species selection and cultivation practices, alongside measures for enhancing wildfire resistance and property protection. All recommendations will aim for implementation within the next 5 to 10 years to align with your timeline.
    \item[Power Grid Manager]
    Step 1: Analyze the Recent fire incident data to understand recent trends and specific areas of higher fire incidence near Covington, VA. This will help in assessing the most vulnerable spots along the transmission lines and around residential areas that need urgent attention for vegetation management and infrastructure protection.\\
    Step 2: Use the Long term fire history records to gather insights into historical wildfire patterns in the area. This can offer a broader context of fire behavior over centuries, helping to predict potential future hotspots and periods of higher risk, which is crucial for long-term planning and resilience strategies.\\
    Step 3: Conduct a literature search focusing on effective strategies for vegetation management, forest health maintenance, and wildfire risk mitigation around power grids. This will include exploring practices such as controlled burns, vegetation clearance standards, and innovations in infrastructure design that enhance resilience to wildfires.\\
    Step 4: Based on the data analysis and best practices identified in scientific literature, provide recommendations tailored to improving forest management and enhancing the resilience of power transmission lines. This will include specific measures to ensure effective clearance around lines, techniques for early fire detection, and community engagement strategies to reduce the overall fire risk.
\end{enumerate}

\subsection*{Information Fidelity of WildfireGPT's Reported Wildfire Statistics}

We verify whether the reported data and citation in WildfireGPT's responses originate from the retrieved source. The fidelity verification process does not require domain expert knowledge but simply needs to match statistics in the responses to those in the retrieved data. We use precision scores to quantify the proportion of correct statistics reported by the language model;  the formula is
\begin{equation}
    \text{Precision} = \frac{\text{Number of correctly reported statistics}}{\text{Total number of statistics reported by the language model}.}
\end{equation} 
In the context of data analysis, if the textual response by WildfireGPT mentions a data point verbatim, then it is considered a correctly reported statistic. For example, consider the following sentence from a WildfireGPT response:

\begin{tcolorbox}
``The wildfire incident data for Mora County from 2015 to 2022 shows a notable increase in fire occurrences, particularly in 2018 with 29 incidents. The most active months for wildfires are during the summer, peaking in July with 21 incidents."
\end{tcolorbox}

In this case we would verify whether the numbers 29 (incidents in 2018) and 21 (incidents in July) are true based on the extracted data. If both numbers match the data, they would be counted as correctly reported statistics. Similarly, in the context of retrieving scientific literature, if WildfireGPT discusses three papers and two of them match the papers returned from the literature retrieval tool, the precision would be $2/3$ or $66.7\%$.

For the literature review, we also calculate the Sentence-BERT (SBERT)~\citep{nils2020sentence} cosine similarity score between the retrieved papers and the generated response of WildfireGPT. The SBERT score ranges from -1 to 1, with higher scores indicating greater semantic similarity. A score above 0.5 is generally considered a good match \citep{reimers2019sentence}. A high SBERT cosine similarity score suggests that the language model is not only citing the correct papers but also capturing the main ideas and concepts from the literature. 

We note that information fidelity between the retrieved sources and WildfireGPT's outputs is not the same as independently validating claims against the broader scientific literature; this stringent view allow us to cleanly verify the model's capacity to accurately synthesize information from credible sources.\footnote{While SBERT is used to find relevant sources by matching queries to our database, we separately use SBERT to check whether the system's final responses actually align with the content of those retrieved sources. This separate evaluation step is necessary since retrieving relevant sources does not guarantee the system will use them appropriately when generating responses. Additionally, while verifying that sources are properly included is a basic requirement (as opposed to hallucinated), it remains an important check for system reliability.}

As shown in Table \ref{tab:correctness_scores}, WildfireGPT achieves high fidelity scores across all case studies (overall success rates of 98.94\% for data analysis and 100\% for literature review, respectively) and high semantic similarity between the papers discussed by the language model and the abstracts of the retrieved papers (0.8559 average SBERT score). This is expected given that our design explicitly incorporates retrieved material, with prompts that specifically direct the language model to accurately summarize the data. Nevertheless, instances of ambiguity occasionally emerge. In the Community Hazard Mitigation Planning case, WildfireGPT stated that "spring FWI projections for the mid-century (2045--2054) and the end of the century (2085--2094) show a gradual increase, particularly notable during the spring season, where the FWI is projected to rise significantly from medium to potentially high-risk levels." According to the retrieved data, however, the projections for the spring seasons were classified as medium risk for both the mid-century (2045--2054) and the end of the century (2085--2094), despite showing an increasing trend, which led to a relatively lower fidelity score. 
    
\begin{table}[htbp]
    \centering
    \caption{Fidelity scores and SBERT scores for each case study. The \edit{fidelity} scores are represented as precision percentages, calculated as the number of correctly reported statistics divided by the total number of statistics reported by WildfireGPT. The raw counts of correctly reported statistics and total statistics are provided. The SBERT scores are reported as average values across all retrieved papers for each case study, along with the number of times papers were retrieved (in parentheses). "N/A" indicates that the case study did not involve a literature review component. Case studies are listed in alphabetical order.\\}
    \label{tab:correctness_scores}
    \begin{tabular}{|p{0.4\linewidth}|c|c|c|}
    \hline
    \textbf{Case Study} & \textbf{Data Analysis} & \textbf{Literature Review} & \textbf{Avg. SBERT Score} \\
    \hline
    Climate Impact on U.S. Wildfires & 5/5 & N/A & N/A \\
    \hline
    Hazard Mitigation Planning & 10/11 & N/A & N/A \\
    \hline
    Comprehensive Wildfire Impact & 16/16 & 5/5 & 0.8529 (2) \\
    \hline
    Current Wildfire Risk Analysis & 11/11 & N/A & N/A \\
    \hline
    Ecosystem Fire Management & 6/6 & 3/3 & 0.9150 (1) \\
    \hline
    Infrastructure Wildfire Risk & 10/10 & 3/3 & 0.8398 (1) \\
    \hline
    Post-Wildfire Public Safety & 1/1 & 3/3 & 0.7742 (1) \\
    \hline
    Private Property Protection & 1/1 & 17/17 & 0.8694 (6) \\
    \hline
    Urban Wildfire Mitigation & 8/8 & 9/9 & 0.8556 (3) \\
    \hline
    Wildland Urban Interface Impact & 12/12 & 3/3 & 0.8844 (1) \\
    \hline
    \textbf{Average Across All Percentages or Scores} & 98.09\% & 100.00\% & 0.8559 \\
    \hline
    \textbf{Overall Success Rate} & 98.94\% (80/81) & 100.00\% (45/45) & N/A \\
    \hline
    \end{tabular}
\end{table}

\subsection*{More Qualitative Feedback}

\begin{figure}[hbt!]
\begin{tcolorbox}[title=Case Study: Private Property Protection]
\textbf{Context:} A Virginia homeowner near Covington seeks advice on forest management, focusing on oak and cherry tree health, while mitigating wildfire risks. After analyzing recent fire data, noting an increase in wildfire incidents, especially during spring and fall, WildfireGPT suggests strategies like fire monitoring, fuel management, controlled burns, defensible spaces, and reforestation with fire-resistant species. Oak trees are highlighted for their fire resilience, making them a smart reforestation investment.\\
\textbf{User Query:} Could you please provide a more detailed summary for investing in oak species?\\
\textbf{WildfireGPT Response (Snapshot):}
\begin{itemize}
    \item Carbon Sequestration: As large and long-lived trees, oaks are effective at capturing and storing carbon dioxide, which could potentially qualify landowners for carbon credits under emerging environmental regulations and markets. 
    \item Mixing Species for Resilience: While investing predominantly in oaks, it is advisable to include a mix of other native, fire-resistant species to ensure ecological balance and adaptability to changing climate conditions.
\end{itemize}
\textbf{Expert Comment:} Carbon sequestration is an insightful point that might not immediately come to mind for many. It's valuable to remind people that trees, especially long-lived species like oaks, offer more than just timber value. They can also serve as carbon sinks, potentially providing landowners with additional income through carbon credits in emerging environmental markets. Mixing species for resilience is also a crucial aspect to highlight. While the focus is on oak species, it's important to emphasize that a diverse planting strategy is more beneficial. Encouraging a mix of native, fire-resistant species alongside oaks promotes greater ecological resilience and adaptability to changing climate conditions. This approach aligns with best practices in forest management and conservation. Both points add depth to the investment strategy, considering both economic and ecological factors.
\end{tcolorbox}
\caption{Private Property Protection: WildfireGPT showcases its ability to provide tailored advice to individual property owners seeking to balance wildfire risk mitigation with their specific land management goals. }
\label{box:Private Property Protection}
\end{figure}

The Private Property Protection case study (Supplementary Figure \ref{box:Private Property Protection}) highlights WildfireGPT's ability to provide insightful ideas that might not be immediately apparent to users. When a domain expert, assuming the role of a Virginia homeowner, sought advice on managing  oak and cherry trees while reducing wildfire risks, WildfireGPT not only suggested standard strategies such as fire monitoring and defensible spaces but also raised an interesting point about the potential for landowners to benefit from carbon credits by investing in oaks. Additionally, WildfireGPT emphasized the importance of mixing oak species with other native, fire-resistant trees to promote ecological balance and adaptability. The domain expert praised these points, noting that they might not immediately come to mind for many but offer valuable economic and ecological benefits. 

\begin{figure}[hbt!]
\begin{tcolorbox}[title=Case Study: Infrastructure Wildfire Risk]
\label{box:Infrastructure Wildfire Risk}
\textbf{Context:} A risk and resilience analyst discussed strategies for mitigating wildfire risks with WildfireGPT, particularly focusing on the use of FWI projections and firebreaks. The conversation has detailed the implementation of firebreaks to protect transmission lines, including site assessment, types, and maintenance of firebreaks. They explored integrating these with other mitigation measures and engaging with local communities for effective implementation.\\
\textbf{User Query:} What goes in to being a critical area as an high-risk areas prone to wildfires. Can you explain?

\textbf{WildfireGPT Response (Snapshot):}\\
\textit{Ecological and Cultural Significance}
\begin{itemize}
    \item Protected Areas: Wildlife reserves, national parks, and other protected lands are often considered high-risk areas due to their ecological importance.
    \item Cultural Heritage: Areas with cultural or historical significance may also be prioritized for protection against wildfires. 
\end{itemize}

\textbf{Expert Comment:} The inclusion of cultural significance is a crucial aspect in explaining the impact of natural disasters, particularly wildfires. This intangible factor is often overlooked in decision-making processes for natural hazard mitigation, yet it remains highly important. Highlighting cultural significance reminds users of the broader, non-quantifiable impacts that wildfires can have on communities and their heritage. It's good that the answer recognizes this often undervalued dimension of wildfire risk assessment.
\end{tcolorbox}
\caption{Infrastructure Wildfire Risk: WildfireGPT demonstrates its ability to provide fresh perspectives and consider often-overlooked dimensions of wildfire risk assessment. }
\label{box:Infrastructure Wildfire Risk}
\end{figure}

Similarly, in the Infrastructure Wildfire Risk case study (Supplementary Figure \ref{box:Infrastructure Wildfire Risk}), a risk and resilience analyst discussed strategies for mitigating wildfire risks with WildfireGPT, focusing on the use of FWI projections and firebreaks. When asked about the factors that contribute to an area being considered high risk for wildfires, WildfireGPT not only mentioned ecological importance but also highlighted the often-overlooked aspect of cultural significance. The domain expert commended the system's recognition of this intangible factor, noting that it reminds users of the broader, non-quantifiable impacts that wildfires can have on communities and their heritage.

\begin{figure}[hbt!]
\begin{tcolorbox}[title=Case Study: Hazard Mitigation Planning]
\textbf{Context:} A hazard mitigation planner for Mora County, NM, is assessing wildfire risks over the next 15 years. In the previous discussion, WildfireGPT found a notable increase in fire occurrences from 2015 to 2022 and a clear trend of increasing wildfire risk in the coming decades due to climate change impacts. From the U.S. census data, the community, with a population of about 10,297, is particularly vulnerable due to high poverty levels. \\
\textbf{User Query:} Let's explore specific adaptation strategies tailored to these demographics (in Mora County, NM).)
\textbf{WildfireGPT Response (Snapshot):}\\
\textit{Enhanced Emergency Communication Systems:}
\begin{itemize}
    \item Develop a multilingual alert system that provides real-time updates on wildfires via SMS, social media, and local radio stations to ensure that all community members receive timely information.
    \item Set up community billboards in key areas to display emergency information and tips on wildfire preparedness.
\end{itemize}

\textbf{Expert Comment:} 
The suggestion to develop a multilingual alert system is insightful, particularly given the significant Spanish-speaking population in the area. Interestingly, this demographic detail wasn't explicitly mentioned in the provided census data, which only included poverty rates and housing units. This indicates that the AI model is drawing on broader knowledge about the region to make this recommendation.
It's noteworthy that the model suggests community-focused strategies rather than individual-based ones. This approach is particularly relevant for a lower-income community like Mora County, where collective efforts may be more effective and feasible for wildfire mitigation.
The model's ability to integrate both explicitly provided information and implicit knowledge about the area demonstrates its capacity for nuanced and context-appropriate recommendations.
\end{tcolorbox}
\caption{Community Hazard Mitigation Planning: WildfireGPT demonstrates its ability to generate nuanced and context-specific recommendations by integrating both explicitly discussed information and implicit knowledge about the unique needs of the target community.}
\label{box:Community Hazard Mitigation Planning}
\end{figure}

The Community Hazard Mitigation Planning case study (Supplementary Figure \ref{box:Community Hazard Mitigation Planning}) further demonstrates WildfireGPT's ability to generate nuanced and context-specific recommendations by integrating both explicitly discussed information and implicit knowledge about the unique needs of the target community. When a hazard mitigation planner for Mora County, NM, sought guidance on wildfire risk adaptation strategies tailored to the community's demographics, WildfireGPT suggested developing a multilingual alert system and focusing on community-based approaches. Despite the census data only including poverty rates and housing units, the system drew on broader knowledge to generate these insightful recommendations, indicating its awareness of the significant Spanish-speaking population in the region. This case study showcases WildfireGPT's potential to offer nuanced decision support that extends beyond mere brainstorming, ultimately enhancing the effectiveness of wildfire risk adaptation efforts.

These case studies collectively underscore WildfireGPT's value in providing insightful, context-specific recommendations that consider the multifaceted nature of wildfire risk management. By integrating diverse information sources and accounting for the ecological, cultural, and socioeconomic dimensions of the communities it serves, the system demonstrates its ability to offer tailored guidance that can inform more effective and feasible wildfire risk mitigation strategies. These case studies highlight WildfireGPT's potential to stimulate innovative thinking and support informed decision-making by generating recommendations that extend beyond conventional approaches, ultimately enhancing the effectiveness of wildfire risk adaptation efforts across various contexts.